\documentclass[letterpaper,journal]{IEEEtran}
\usepackage{amsmath,amsfonts}
\usepackage{algorithmic}
\usepackage{algorithm}
\usepackage{array}
\usepackage[caption=false,font=normalsize,labelfont=sf,textfont=sf]{subfig}
\usepackage{textcomp}
\usepackage{stfloats} 
\usepackage{url}
\usepackage{verbatim}
\usepackage{graphicx}
\usepackage{cite}
\usepackage{multirow}
\usepackage{soul, color, xcolor}
\makeatletter

\newcommand{\Rmnum}[1]{\expandafter\@slowromancap\romannumeral #1@}
\makeatother

\usepackage{tikz,xcolor,hyperref}
\definecolor{lime}{HTML}{A6CE39}
\DeclareRobustCommand{\orcidicon}{%
\begin{tikzpicture}
\draw[lime, fill=lime] (0,0)
circle [radius=0.16]
node[white] {{\fontfamily{qag}\selectfont \tiny ID}};    \draw[white, fill=white] (-0.0625,0.095)
circle [radius=0.007];    \end{tikzpicture}
\hspace{-2mm}}
\foreach \x in {A, ..., Z}{%
\expandafter\xdef\csname orcid\x\endcsname{\noexpand\href{https://orcid.org/\csname orcidauthor\x\endcsname}{\noexpand\orcidicon}}
}

\begin{document}

\title{QDS-SNN: Energy-efficient Quantum Deeply-Supervised Spiking Neural Network Algorithm for Traffic Sign Recognition}

\author{Zhiguo Qu, Keqi Li, Le Sun, Wenjie Liu, Yimin Yu, Saif Al-Kuwari and Ahmed Farouk
\thanks{This work was supported by the National Natural Science Foundation of China (62071240, 62462065, 72441009), Major Science and Technology Special Project of Yunnan Province (202402AD080005) and Special Project for the Construction of the Science and Technology Innovation Center for South and Southeast Asia (202203AP140010), funded by the Yunnan Provincial Science and Technology Department. It also is supported by in part by Quantum Science and Technology-National Science and Technology Major Project under Grant 2021ZD0302901.}
\thanks{Zhiguo Qu, Le Sun and Wenjie Liu are with the School of Computer Science, School of Software, Nanjing University of Information Science and Technology, Nanjing 210044, China (e-mail: 002359@nuist.edu.cn, 002813@nuist.edu.cn, 001281@nuist.edu.cn).}
\thanks{Keqi Li is with the School of Computer Science, School of Software, Nanjing University of Information Science and Technology, Nanjing 210044, China (e-mail: 202312490521@nuist.edu.cn).}
\thanks{Yimin Yu is with the School of Information, Yunnan University of Finance and Economics, Kunming 650221, China (e-mail: yym@ynufe.edu.cn).}
\thanks{Saif Al-Kuwari is with Qatar Center for Quantum Computing, College of Science and Engineering, Hamad Bin Khalifa University, Doha, Qatar (e-mail: smalkuwari@hbku.edu.qa).}
\thanks{Ahmed Farouk is with Qatar Center for Quantum Computing, College of Science and Engineering, Hamad Bin Khalifa University, Doha, Qatar, and the Department of Computer Science, Faculty of Computers and Artificial Intelligence, Hurghada University, Hurghada, Egypt (e-mail: ahmedfarouk@ieee.org).}
}

\markboth{Internet of Things Journal,~Vol.~, No.~, ~}%
{Shell \MakeLowercase{\textit{et al.}}: QDS-SNN: Energy-efficient Quantum Deeply-Supervised Spiking Neural Network Algorithm for Traffic Sign Recognition}

\maketitle

\begin{abstract}
Traffic sign recognition is crucial for intelligent transportation and autonomous driving, as it can improve driving efficiency and ensure road safety. However, traditional recognition methods are based on large datasets and intensive computation, which limits their real-time applicability. Spiking Neural Networks (SNNs) offer a biologically inspired, energy-efficient alternative due to their spatiotemporal processing capabilities, but suffer from information loss and vanishing gradients during training. To overcome these limitations, this study proposes a Quantum Deep-supervised Spiking Neural Network (QDS-SNN) that integrates Quantum Neural Networks (QNNs) for efficient, low-power deep supervision. Using quantum superposition and entanglement, QNNs enable expressive representations and parallel computation, thereby enhancing performance without compromising energy efficiency. The proposed QDS-SNN incorporates a temporally and spatially adaptive LIF (TSA-LIF) neuron and a quantum-assisted classifier module (QACM) to mitigate gradient issues and improve training effectiveness. This study conducts experiments on the PennyLane quantum simulation platform, and the results show that QDS-SNN achieves 99.72\% accuracy on the GTSRB dataset in only 6 time steps—outperforming the MS-ResNet baseline by 1.32\% while reducing energy consumption by 55.77\%. In the TSRD dataset, it achieves 97.90\% accuracy while reducing energy use to 52.68\% of the baseline. These results demonstrate that QDS-SNN offers a high-performance, energy-efficient solution for traffic sign recognition in intelligent transportation systems.
\end{abstract}


\section{Introduction}
\IEEEPARstart {W}{ith} the advancement of the Internet of Vehicles (IoV), traffic sign recognition is indispensable for autonomous driving, improving safety and efficiency. Traffic signs convey road conditions, hazards, and regulatory information through standardized colors and symbols to guide vehicle operations and driving decisions in real time. Accurate recognition is therefore critical for effective driver-assistance and safe autonomous navigation.

Convolutional neural networks (CNNs) \cite{ref1} have long been the standard for traffic sign recognition due to their high accuracy. However, they have obvious limitations: reliance on large-scale labeled datasets and on massive computational resources that increase sharply with task complexity, and insufficient parallel processing capacity for complex tasks. In addition, their high computational overhead results in significant energy consumption, making them unsuitable for resource-constrained, energy-sensitive applications such as electric vehicles. Thus, there is an urgent need to explore alternative approaches that balance performance with high efficiency and low power consumption to meet the demands of green computing.

QNNs have advanced rapidly in recent years, leveraging quantum parallelism and superposition to tackle complex problems and achieve superior performance over traditional architectures in intelligent transportation, smart healthcare, and other fields \cite{ref2}. When combined with Spiking Neural Networks (SNNs), QNNs enable simultaneous analysis of temporal features across multiple time steps, thereby improving the efficiency of dynamic image recognition and driving the development of more efficient, low-power neural models for broader computer vision applications. However, integrating QNNs with SNNs poses notable challenges: their inherently different computational paradigms, along with QNNs' high sensitivity to quantum noise, necessitate effective noise mitigation to ensure the accuracy and stability of hybrid QNN-SNN models.

In the IoV environment, traffic sign recognition systems operate under limited computational resources and strict energy constraints. Although deep CNN-based methods achieve high accuracy, their computational complexity and energy consumption hinder large-scale deployment in IoV scenarios \cite{ref_ITSRS}. Therefore, improving model efficiency while reducing energy consumption is crucial for practical applications. To address this, this study proposes an energy-efficient traffic sign recognition algorithm based on a QDS-SNN. This approach integrates a QACM and a TSA-LIF neuron to improve performance and reduce energy consumption. Additionally, a hybrid training strategy leverages the distinct advantages of SNNs and QNNs by employing a joint loss function to optimize both networks simultaneously, thereby accelerating convergence and improving training efficiency. In general, QDS-SNN achieves accurate and energy-efficient traffic sign recognition on the pennylane quantum simulation platform, with significant potential for deployment in intelligent transportation systems. The main contributions of this work are as follows:
\begin{itemize}
\item[1)] This study proposes the QDS-SNN algorithm, which integrates QACM and TSA-LIF into a K-stage deep SNN architecture. This algorithm demonstrates higher performance and lower energy consumption when processing traffic sign classification tasks.

\item[2)] This study designs the TSA-LIF neuron model with learnable temporal and spatial decay factors. The spatial factor controls feature retention, while the temporal factor regulates memory decay, thereby improving the network's temporal modeling capability.

\item[3)] This study develops a Quantum-assisted Classification Module (QACM) that takes advantage of low-power quantum computing to enhance deep supervision. QACM improves performance while reducing energy consumption, avoiding the overhead of traditional convolution.

\item[4)] This study introduces a hybrid training method that optimizes QNN and SNN jointly, using a combined loss function with a supervision factor to balance deep supervision. Continuous loss minimization accelerates convergence and improves performance.

\end{itemize}

The remainder of this paper is organized as follows. Section II reviews the related work. Section III introduces the proposed QDS-SNN model in detail. Section IV provides experimental results and analysis. Finally, the conclusions are given in Section V.

\section{Related Work}
\subsection{Traffic sign recognition}
Traffic sign recognition (TSR) plays an important role in intelligent transportation systems by providing real-time road information for perception and decision-making in autonomous vehicles. It contributes to improved safety and efficiency and supports the development of automated and green transportation. Most TSR approaches rely on convolutional neural networks (CNNs) for feature extraction and classification. Through convolution and pooling operations, CNNs learn hierarchical representations that effectively capture local patterns and visual structures, making them highly successful in image recognition tasks.

Research on traffic sign recognition has evolved significantly. With advances in computing, Shustanov et al. \cite{ref6} introduced CNNs to this task in 2017, significantly improving classification performance in complex road environments. In 2018, Hu et al. \cite{ref7} applied SENet to model inter-channel dependencies, improving feature representation through selective attention. In 2020, Hasan et al. \cite{ref8} combined SVM and CNN to balance classification accuracy and discrimination across 12 sign categories. In 2023, Lian et al. \cite{ref9} integrated federated learning and model sparsification into an IoV-based TSR and adopted the Adam optimizer for local training to improve recognition performance, ensure data privacy and reduce communication overhead. In 2024, You et al. \cite{ref10} proposed SLMFed, a stage-layerwise incremental federated learning mechanism for GTSRB traffic sign recognition, which increases accuracy and reduces communication costs in dynamic IoT scenarios. Recent efforts have explored quantum computing to increase TSR efficiency. In 2025, Qu et al. \cite{ref2} proposed QCACNN, which includes a Quantum Channel Attention Layer and achieves competitive accuracy at a significantly lower computational cost, demonstrating promise for resource-constrained scenarios.

CNNs perform well, but require significant computational power and energy, limiting their use on resource-constrained devices, such as autonomous vehicles. SNNs, inspired by biological event-driven processing, offer low power consumption while maintaining accuracy, making them well-suited for energy-efficient intelligent transportation.
 Recent research on SNN-based TSR has made notable progress. In 2022, Xie et al. \cite{ref10_1} proposed a FedSNN-NRFE scheme based on neuronal receptive field encoding, achieving high-precision TSR with superior energy efficiency in IoV scenarios. In 2023, Zhang et al. \cite{ref10_4} designed a hybrid SNN-CNN network using RRAM weights, achieving reliable recognition accuracy while reducing energy consumption. In 2024, Chen et al. \cite{ref10_2} designed an SA-SCNN model incorporating a spatial attention mechanism, achieving 99.56\% accuracy on the GTSRB dataset. In 2025, Yadav et al. \cite{ref10_3} proposed an energy-efficient CSNN with energy consumption only 4.8\% of that of an equivalent CNN, achieving 99.90\% accuracy on the GTSRB.

In summary, CNNs are the main TSR method with remarkable progress, but their high computational and energy consumption limits their application in resource-constrained scenarios. SNNs, because of their low-power advantages, have become a promising alternative and have achieved notable progress in recent TSR research.

\subsection{Spiking neural network}
CNNs enable accurate processing of vehicle data, but are computationally intensive and energy-hungry, limiting their use in resource-constrained IoV scenarios. SNNs, inspired by biology, offer low-power, fast, event-driven alternatives with spatiotemporal dynamics and binary spikes \cite{ref11}. However, their non-differentiable activations hinder gradient-based training. There are two main approaches: convert ANNs to SNNs \cite{ref12}, which requires long simulations and high energy, and direct training with surrogate gradients.

Early SNN research in IoV lacked practical adaptability to complex TSR scenarios. In 2016, Kaiser et al. \cite{ref13} proposed a retina-inspired visual encoding for SNNs for lane following, but this model was limited to simple lane control and failed to meet the high-precision recognition needs of various applications. In 2020, Bing et al. \cite{ref14} combined R-STDP with reinforcement learning for end-to-end IoV training, yet their reinforcement learning-based approach suffered from slow convergence and poor robustness in dynamic TSR environments. Subsequent work focused on training efficiency but still had critical gaps: In 2021, Chandarana et al. \cite{ref15} proposed adaptive static image encoding for low-power edge computing, but their encoding method ignored the temporal dynamics of moving traffic signs, degrading recognition accuracy. In the same year, Kim et al. \cite{ref16} developed time-batch normalization for high-precision SNNs, but this technique increased computational overhead, undermining SNNs’ inherent low-energy advantage. In the same year, Zheng et al. \cite{ref17} proposed STBP-tdBN for direct deep SNN training, yet their method struggled with gradient instability in deep architectures to learn complex TSR features. In 2022, Deng et al. \cite{ref18} optimized convergence using the TET method, but this only improved training speed without enhancing the model's adaptability to TSR’s dynamic scenarios. More recently, in 2024, Chen et al. \cite{ref19} designed the SA-SCNN TSR model with spatial attention, but its reliance on the classical SNN architecture failed to break the accuracy-energy tradeoff for embedded IoV devices. In 2024, Zuo et al. \cite{ref_Shrink} explored the trade-off between time-step reduction and network-layer design for low-latency neuromorphic object recognition, providing insights into SNN structure optimization. In 2025, Yu et al. \cite{ref20} proposed a time-shift module to fuse multi-temporal spike features, yet this module added complexity without leveraging quantum advantages to further reduce energy consumption—a critical consideration for resource-constrained TSR deployment. In the same year, Yadav et al. \cite{ref_yadav} proposed a convolutional spiking neural network for traffic sign recognition, which achieved a balance between network structure optimization and energy efficiency. However, this method still relies on the traditional spiking architecture and is difficult to meet the needs of complex scenarios. In the same year, Shen et al. \cite{ref_Deploy} analyzed the joint impact of layer simplification and time-step optimization on SNN performance in real-world intelligent transportation scenarios, enriching research on the practical deployment of related models.

It should be noted that while SNNs excel in energy-efficient TSR due to their sparse, event-driven computation, they lack the ability to process the high-dimensional features required for complex traffic signs. Therefore, integrating quantum computing with SNNs is a promising approach. It leverages the parallelism and superposition properties of quantum computing to accelerate computation while retaining the spiking mechanism of SNNs, creating a synergistic framework that addresses the accuracy-energy tradeoff that standalone SNNs or CNNs cannot satisfy in IoV applications.

\subsection{Quantum neural network}
QNNs use qubits and quantum gates for computation and learning. The key difference between them and classical neural networks lies in their use of quantum superposition and entanglement to improve computational efficiency and enable more powerful information processing. Quantum advantage has great potential to overcome the bottleneck in classical neural networks for high-dimensional feature processing.

QNNs originated from early studies on quantum neural computing, such as the work of Kak et al. \cite{ref21}. Since then, the field has developed substantially. For example, Wiebe et al. \cite{ref24} represented classical vectors as amplitudes of quantum states and proposed a quantum nearest-neighbor method to compare distances between quantum states.

In the Noisy Intermediate-Scale Quantum (NISQ) era \cite{ref26}, circuit-constructed QNNs and hybrid quantum-classical algorithms have attracted increasing attention, leveraging configurable quantum gates to enable task-specific tuning and offering advantages over classical counterparts. Yet such works suffer from a narrow focus on applications and poor adaptation to TSR's core demands for high precision and low energy consumption. In 2018, Farhi et al. \cite{ref27} proposed a QNN framework for binary classification, laying the foundation for quantum-assisted classification, but it was limited to simple binary tasks and was unsuitable for TSR's complex multi-class and real-time inference needs. In 2020, Schuld et al. \cite{ref28} optimized this architecture with a circuit-focused quantum-assisted classifier, simplifying QNN construction but achieving only basic classification and failing to mitigate gradient vanishing during training of complex deep networks. In 2024, Qu et al. \cite{ref29} integrated quantum federated learning (QFL) with a quantum minimal gated unit (QMGU) into 5G IoV, proposing the QFSM algorithm to enhance computational efficiency and privacy protection. In 2025, Qu et al. \cite{ref30} proposed DAQFL, a dynamic aggregation QFL algorithm for intelligent diagnosis in the Internet of Medical Things, thereby further expanding the applications of quantum federated learning in resource-constrained IoT scenarios.

Quantum Spiking Neural Networks (QSNNs) integrate quantum computing with SNNs but remain ill-suited for TSR tasks. In 2024, Brand et al. \cite{ref31} proposed a quantum leaky integrate-and-fire neuron that outperforms classical SNNs on MNIST but lacks deep supervision, thereby limiting feature learning in dynamic TSR scenarios. 
In 2025, Jha et al. \cite{ref_jha} proposed a hybrid SNN-quantum framework for spatiotemporal data classification, utilizing SNN-derived spike frequency features and quantum kernels to enhance classification performance; however, this framework was designed for EEG data and is difficult to apply to visual feature extraction.
In the same year, Liu et al.\cite{ref32} developed a cost-efficient QSNN with fewer parameters, yet it neglects energy efficiency—critical for onboard deployment. Fundamentally, existing QSNNs achieve only superficial integration of quantum components and SNNs, lacking a task-specific, synergistic design for TSR.

In summary, previous research on QNNs and QSNNs suffers from key shortcomings: insufficient integration of deep supervision, failure to meet the energy-efficiency requirements of TSR tasks, and limited performance. Our QDS-SNN employs quantum deep supervision to mitigate the vanishing-gradient problem and enhance robustness; it combines QACM with TSA-LIF neurons and proposes a framework for resource-constrained intelligent transportation. This deep, task-oriented integration is unique to our work and provides a practical, high-performance solution for intelligent transportation.

\begin{figure*}
    \centering
    \includegraphics[width=0.7\linewidth]{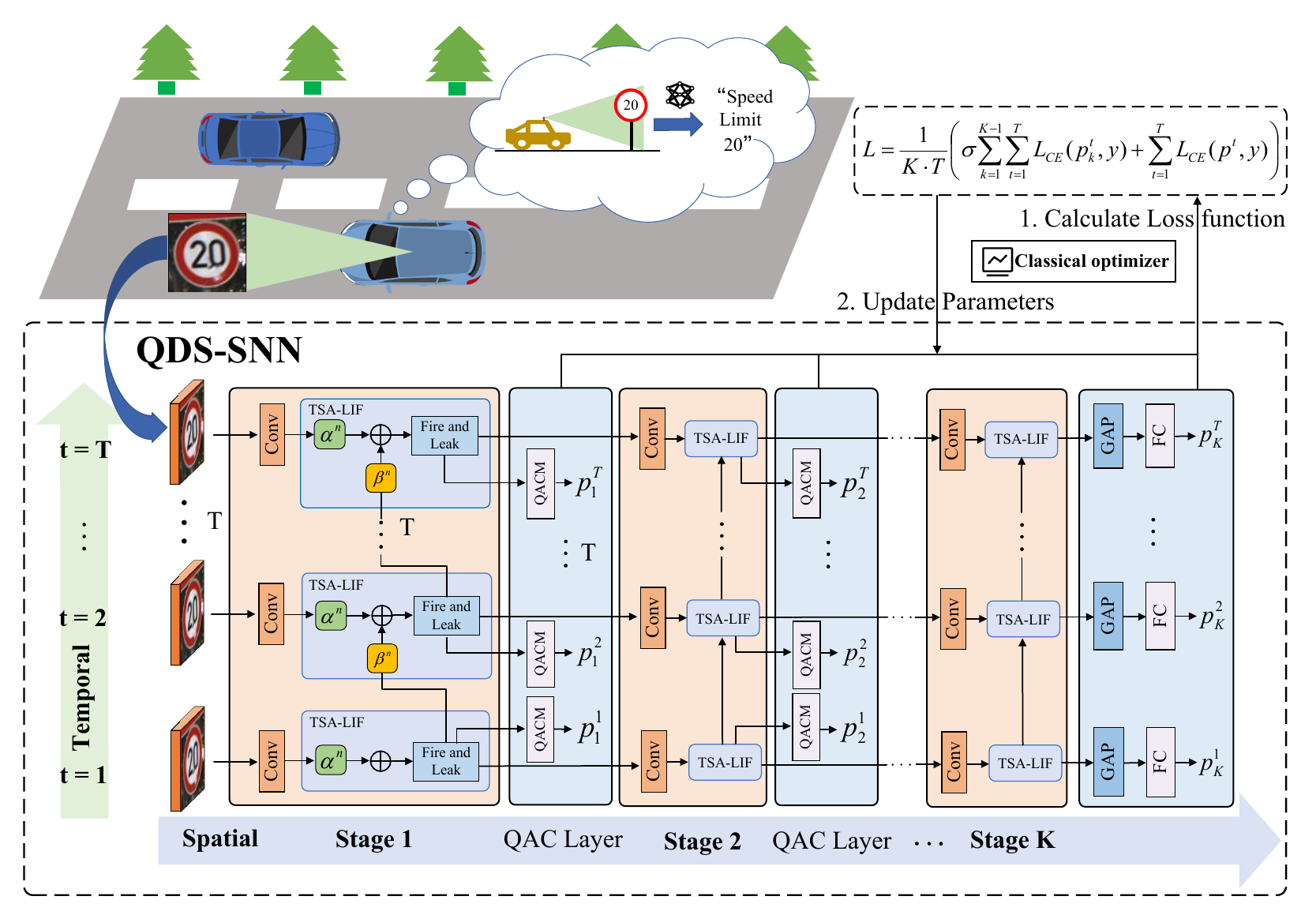}
    \caption{The Proposed QDS-SNN Framework.}
    \label{fig:1}
\end{figure*}

\section{QDS-SNN}
\subsection{Overall framework}
This section presents the proposed QDS-SNN algorithm, which integrates the Quantum-Assisted Classification Module (QACM) and the Spatio-Temporal Adaptive LIF neuron (TSALIF) into a K-stage deep SNN framework.

Deep SNNs commonly suffer from vanishing gradients due to long temporal dependencies and non-differentiable activation functions. To address this issue, QACM is added after each stage to enhance gradient propagation through quantum convolution, pooling, and measurement. Traditional SNN training fails because the Heaviside function is nondifferentiable, preventing direct backpropagation. Surrogate-gradient methods enable optimization but remain prone to gradient vanishing. Therefore, this study proposes TSA-LIF neurons and a hybrid QNN-SNN training method to improve convergence and efficiency while reducing computational and energy overhead. In IoV settings, onboard cameras capture real-time images of traffic signs, and the trained QDS-SNN performs fast, energy-efficient recognition to provide timely driving guidance, thereby improving road safety and the driving experience.

\subsection{Temporal-Spatial Adaptive-LIF (TSA-LIF)}
\subsubsection{Classic LIF neuron model}
Spiking neurons act as the main computational units in SNNs. Among various neuron models, the LIF model is commonly used in deep SNNs. To adapt the LIF model for deep learning, it must be discretized into an iterative equation, as shown below.

\begin{equation}
\label{eq2}
V^{t,n} = f\left( H^{t-1,n}, X^{t,n} \right), 
\end{equation}
\begin{equation}
\label{eq3}
S^{t,n} = \Theta\left( V^{t,n} - V_{\text{th}} \right), 
\end{equation}
\begin{equation}
\label{eq4}
H^{t,n} = V_{\text{reset}} \cdot S^{t,n} + V^{t,n} \odot \left( 1 - S^{t,n} \right)
\end{equation}
Where $V^{t,n}$ is the integrated membrane potential, $H^{t-1,n}$ represents the membrane potential after the spike is activated at the previous time step. The input of the n-th layer at the $t$-th time step $X^{t,n}$ is calculated by the output of the previous layer $S^{t,n-1}$ through a convolutional layer with weight $W$:
\begin{equation}
\label{eq5}
X^{t,n} = \mathrm{Conv}(W, S^{t,n-1}).
\end{equation}

The function of the classic LIF model $f(\cdot)$ is expressed as:
\begin{equation}
\label{eq6}
\mathrm{V^{t,n}=H^{t-1,n}+\frac{1}{\tau_m}(X^{t,n}+(-(H^{t-1,n}-V_{rest}))).}
\end{equation}

The Heaviside function $H(x)$ returns 0 when $x < 0$ and 1 otherwise. The membrane potential $H^{t-1,n}$ is then reset to $V_{reset}$ when $S^{t,n}=1$, and remains $V^{t,n}$ otherwise. In this study, $V_{rest}$ and $V_{reset}$ are set to 0.

\subsubsection{Temporal-Spatial Adaptive-LIF neuron model}
\begin{figure}
    \centering
    \includegraphics[width=1\linewidth]{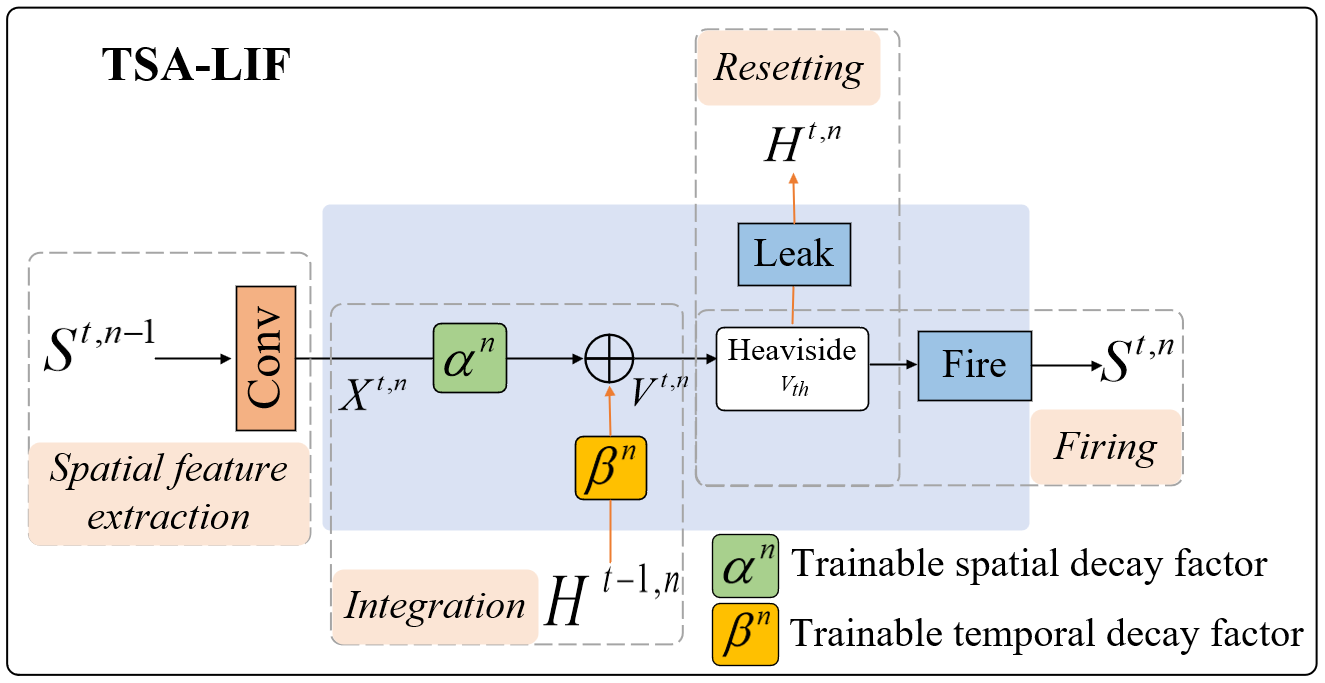}
    \caption{Temporal-Spatial Adaptive-LIF (TSA-LIF).}
    \label{fig:2}
\end{figure}

To enhance adaptability, this study introduces variable attenuation terms ($\mu$ and $\lambda$) into the LIF model, yielding the Temporal-Spatial Adaptive LIF neuron (TSA-LIF), with design insights drawn from the DA-LIF model \cite {ref_DALIF}.
TSA-LIF incorporates trainable spatial and temporal attenuation factors: the spatial factor modulates current input features to control spatial retention and decay, while the temporal factor adjusts the previous membrane potential to regulate forgetting of historical states. This allows for independent decay of the membrane potential and input signals, thus improving temporal modeling.

\begin{equation}
\label{eq7}
\mathrm{\tau_m\frac{du(t)}{dt}=-\mu(u(t)-u_{rest})+\lambda I(t).}
\end{equation}

Then, the integration of the neuron model with variable decay can be expressed as:
\begin{equation}
\label{eq8}
\mathrm{V^{t,n}=(1-\frac{\mu}{\tau_m})H^{t-1,n}+\frac{\lambda}{\tau_m}X^{t,n}.}
\end{equation}

Independent learning of the temporal and spatial decay parameters is distributed using $1-\frac{\mu}{\tau_{\mathrm{m}}}$ and $\frac{\lambda}{\tau_{\mathrm{m}}}$ as different variable decay terms. First, the spatial decay factor $\alpha^n$ and the temporal decay factor $\beta^n$ are defined as follows:
\begin{equation}
\label{eq9}
\alpha^\mathrm{n}=\frac{\lambda^\mathrm{n}}{\tau_\mathrm{m}},\beta^\mathrm{n}=1-\frac{\mu^\mathrm{n}}{\tau_\mathrm{m}}.
\end{equation}

Our approach aims to train these two learnable decay factors independently, defining:
\begin{equation}
\label{eq10}
\mathrm{V^{t,n}=\beta^n H^{t-1,n}+\alpha^n X^{t,n}.}
\end{equation}

The spatial and temporal decay factors are learned independently, sharing parameters within the same layer across time steps but varying between different layers.

TSA-LIF achieves fundamental innovation at the core modeling level, supporting the direct encoding of static images and enabling output features to naturally satisfy deep supervision constraints through membrane-potential normalization and controllable soft-reset mechanisms. At the same time, TSA-LIF is tightly coupled with traffic sign recognition tasks, enabling direct processing of raw traffic sign images without complex preprocessing. The output features are seamlessly integrated with the deep supervision mechanism, significantly reducing computational overhead while ensuring recognition accuracy. This makes the entire model better suited to the real-time, efficient, and high-precision requirements of traffic sign recognition.

\subsection{Quantum-Assisted Classification Module (QACM)}
\begin{figure}
    \centering
    \includegraphics[width=1\linewidth]{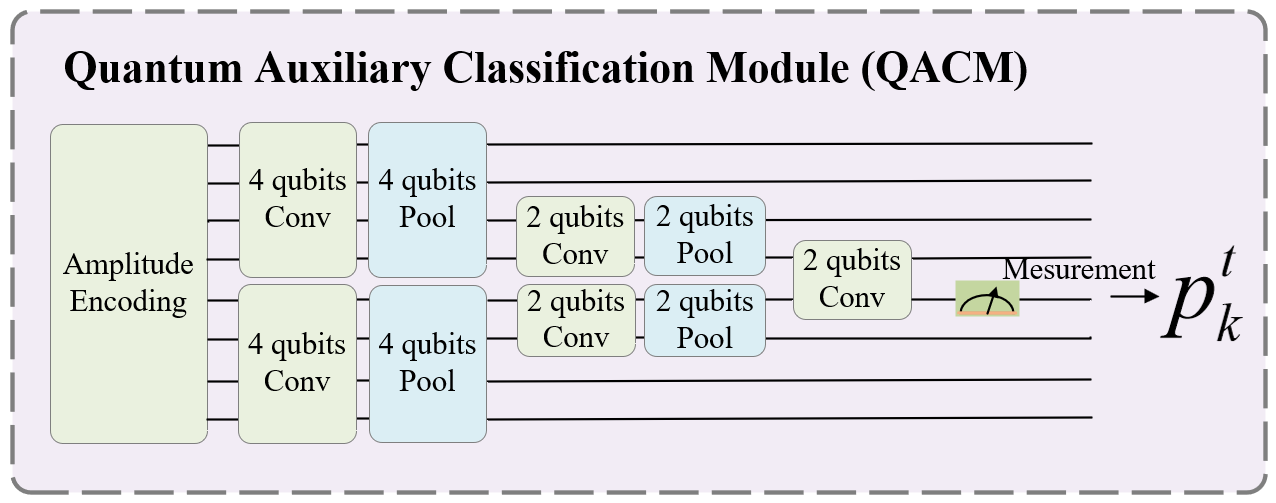}
    \caption{Quantum-Assisted Classification Module (QACM).}
    \label{fig:3}
\end{figure}

In our proposed quantum pulse neural network architecture for traffic sign recognition, a quantum-assisted classification layer (QAC Layer) is added after each stage. This layer employs the quantum-assisted classification module (QACM) to provide deep spatiotemporal supervision, enabling deep fusion of classical and quantum features. As shown in Fig. \ref{fig:3}, the QACM consists of a quantum coding layer, a quantum convolution layer (4-qubit and 2-qubit variants), a quantum pooling layer (4-qubit and 2-qubit), and a measurement layer. 

\textbf{Quantum Coding Layer:} Classical pulse data cannot be directly used to train quantum circuits and must be encoded into quantum states via a quantum coding layer. Inputs to this layer come from the convolutional output and TSA-LIF neurons. For each time step $t$, the features extracted from the LIF neurons are recorded as $x^t \in R^N$, where $N$ represents the flattened spatial feature dimension. The inherent sparsity and event-driven nature of the SNN capture local dynamic representations of traffic signs at each time step $t$.

Here, the quantum coding scheme adopts the amplitude encoding. The core idea is to normalize the classical data vector and map it to the probability amplitude of the computational ground state of an n-qubit system to form a quantum state $|\psi^{t}\rangle$:
\begin{equation}
\label{eq10a}
|\psi^{t}\rangle=\sum_{\mathrm{i=0}}^{\mathrm{N-1}}{x_i^t}|{i}\rangle\ , i=0,1,2,\cdotp\cdotp\cdotp,\mathrm{N}-1
\end{equation}
Here, $|i\rangle$ represents the computational ground state. $\alpha_i$ is the complex probability amplitude corresponding to the ground state $|i\rangle$. These amplitudes must satisfy the normalization condition: $\sum_{i=0}^{N-1}|x_i^t|^2=1$. This method embeds classical features into the amplitudes of the quantum state of $log_2N$ qubits. It has strong expressive power and low resource overhead, making it particularly suitable for compactly representing high-dimensional traffic-image features with limited qubit resources. 

During encoding, to ensure amplitude normalization, the input vector at each time step must first be standardized. After encoding, the quantum state $|\psi^{t}\rangle$ passes through the quantum convolution layer, the quantum pooling layer, and the measurement layer in sequence, and finally outputs the classification result of the $k$-th stage in the $t$-th time step. This intermediate output is used not only for auxiliary supervision, but also, to a certain extent, to improve the interpretability of training and the efficiency of quantum feature extraction.

\begin{figure}
    \centering
    \includegraphics[width=1\linewidth]{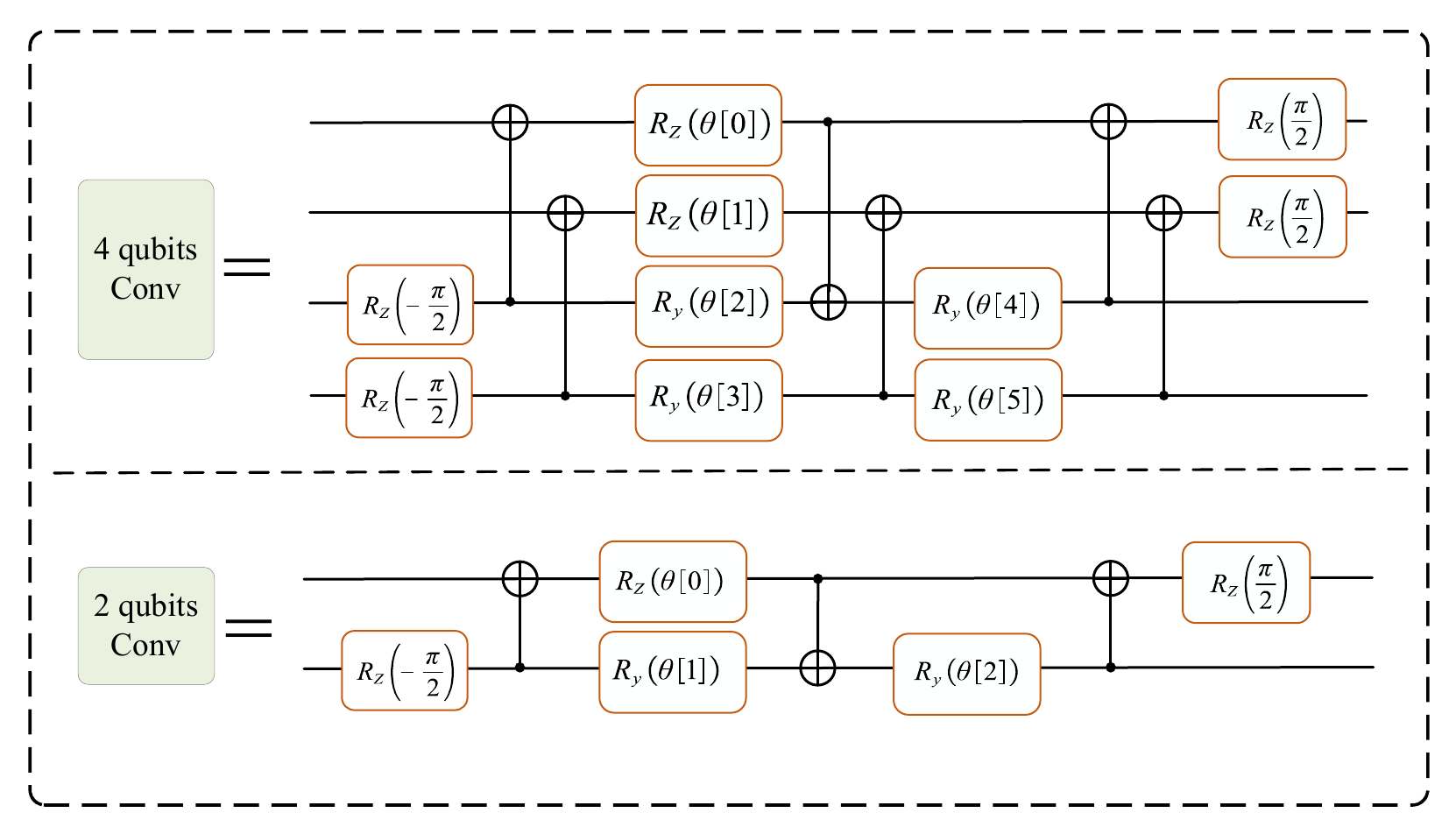}
    \caption{Quantum Convolution Layer Circuit.}
    \label{fig:4}
\end{figure}

\textbf{Quantum Convolution Layer:} It is the core of QACM, leveraging qubit entanglement and parallelism to simulate classical convolution feature extraction and perform nonlinear transformations via trainable parameters. In QACM, convolution layers are categorized by qubit number: 4-qubit and 2-qubit layers, as shown in Fig. \ref{fig:4}.

The 4-qubit convolution layer has two quantum gate layers: first, a fixed Hadamard transform ($R_z(-\frac{\pi}{2})$) distributes the initial state information evenly; second, a parameterized rotation gate array with $R_z(\theta)$, $R_y(\theta)$, and CNOT gates mixes information between qubits. Parallel operation simulates multi-channel outputs, such as CNN kernels. The kernel parameters $\theta[i]$ can be trained by backpropagation. Due to the reversibility of quantum gates, information loss is theoretically avoided, thereby preserving quantum correlations in input features.

The 2-qubit convolutional layer processes compressed quantum features in QACM. It is lightweight, consisting of two CNOT gates and three sets of parameterized rotation gates, reducing gate depth and resource use while preserving quantum expressiveness.

\begin{figure}
    \centering
    \includegraphics[width=1\linewidth]{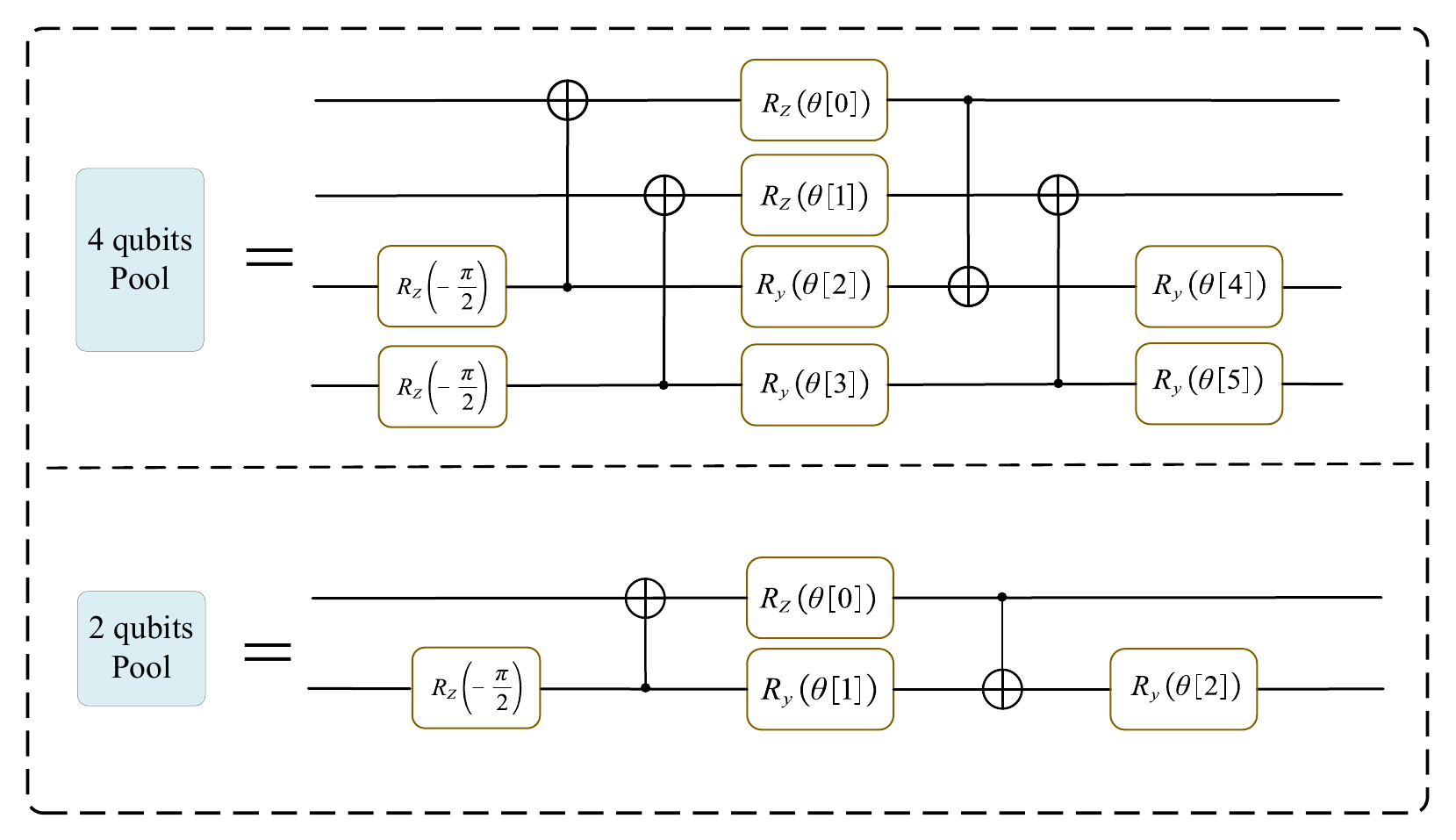}
    \caption{Quantum Pooling Layer Circuit.}
    \label{fig:5}
\end{figure}

\textbf{Quantum Pooling Layer:} To compress the quantum state's dimensions, extract global semantics, and improve generalization, this study introduces a quantum pooling layer after the quantum convolution layer (Fig. \ref{fig:5}). Inspired by classical pooling, it fuses information across qubits via specific quantum gates, reducing the feature space and the number of measurement qubits, lowering computational cost while capturing multi-scale spatiotemporal features in traffic sign images.

The 4-qubit pooling layer compresses information from 4 qubits into 2 target qubits using CNOT and parameterized rotation gates. Rotation gates enable feature aggregation, retaining only a few representative qubits for the next quantum layer.

The 2-qubit pooling layer further compresses qubits while preserving basic feature selection and adapting to the network’s backend. It uses CNOT and rotation gates to merge quantum information.

Through multilevel quantum convolution and pooling, QACM progressively extracts quantum features and produces classification results via measurement. This module improves discrimination and generalization by leveraging the high-dimensional feature space and superposition in quantum computing.

At the end of QACM, the measurement layer converts the quantum information into classical probability distributions for the final classification. After multiple quantum convolution and pooling layers, qubits encode high-dimensional features, and the measurement layer translates these into interpretable classical output while preserving essential information. Assume that the final quantum state processed by the convolution and pooling layers is $|\psi^{t}_{k}\rangle$, and the measurement operation takes the following form:

\begin{equation}
\label{eq13}
\langle Z_k^t\rangle=\langle\psi_k^t| Z|\psi_k^t\rangle
\end{equation}
Here, Z denotes the $Pauli-Z$ measurement operation. The expected values obtained from measurement form a vector of real values representing the result of the traffic sign classification model at the $k$-th stage for time step $t$.

For an SNN with K stages, $\{ {p^t|t\in[1,T]}\}$ denotes the classification output at each time step. As shown in Fig. \ref{fig:1}, to apply deep supervision to the first stages $K-1$, this study introduces QACM after each stage. The QACM quantum circuit consists of a quantum coding layer, a convolution layer, a pooling layer, and a measurement layer. Let $\{f_k(\cdot)\mid k=1,\cdots, K-1\}$ denote the QACM operations and $\{{X_k^t\mid t=1,\cdots,T;k=1,\cdots,K-1}\}$ represent the output feature maps from the first $K-1$ stages and time steps in the SNN. These features are fed to the auxiliary classifier to generate auxiliary predictions:

\begin{equation}
\label{eq14}
p_k^t=f_k(X_k^t)~~ k=1,\cdots,K-1;~t=1,\cdots,T.
\end{equation}

\subsection{QNN and SNN hybrid training method}
During training, a hybrid method that combines SNNs and QNNs is introduced to fully exploit their respective advantages. This method takes into account the characteristics of both networks and designs a hybrid loss function to improve the overall model performance.

For training a deep SNN, this study uses temporal channel backpropagation (STBP) \cite{ref33}. For the training of QNN quantum circuits, the loss function LQACM of the quantum circuit is calculated by the cross-entropy loss function. The auxiliary classification result of QACM is expressed by the expected value of quantum state measurement, and the loss function is expressed as equation \eqref{eq16}:
\begin{equation}
\label{eq16}
L_{QACM}(\theta)=-\frac{1}{N}\sum_iy_i\ln E_i(\theta)+(1-y_i)ln(1-E_i(\theta))
\end{equation}
Here, the expected value $E_{i}(\theta)$ of the quantum state measurement can be measured by equation \ref{eq13}. Next, the gradient of the QNN loss function should be calculated. For the $k$-th parameter to be optimized, the gradient of $\theta_{k}$ is expressed as follows:

\begin{equation}
\label{eq17}
\frac{\partial L_{QACM}(\theta)}{\partial\theta_{k}}=-\frac{1}{N}\sum_{i}\frac{\mathrm{y_{i}}}{E_{i}(\theta)}\frac{\partial E_{i}(\theta)}{\partial\theta_{k}}-\frac{1-y_{i}}{1-E_{i}(\theta)}\frac{\partial E_{i}(\theta)}{\partial\theta_{k}}.
\end{equation}

It is sufficient to compute the partial derivative of $E_i(\theta)$ with respect to the parameter $\theta_k$. Given that the quantum circuits employed here are entirely composed of Pauli rotation gates, this study adopts the differentiation results provided by Schuld et al. \cite{ref34}. Denoting $E_i(\theta)$ as $E$, the derivative of $E$ with respect to $\theta_k$ can be expressed as:
\begin{equation}
\begin{aligned}
\label{eq18}
\frac{\partial\mathrm{E}(\theta)}{\partial\theta_{\mathrm{k}}}=&\frac{\partial\langle\psi^{^{\prime}}|\mathrm{M}|\psi^{^{\prime}}\rangle}{\partial\theta_{\mathrm{k}}}=\frac{\partial\left\langle\psi\left|\mathrm{U}_{\mathrm{a}}^{\dagger}\mathrm{M}\mathrm{U}_{\mathrm{a}}\right|\psi\right\rangle}{\partial\theta_{\mathrm{k}}}\\=&\frac12\Big(\left\langle\psi\left|U_\mathrm{a}^\dagger\left(\theta_\mathrm{k}+\frac\pi2\right)MU_\mathrm{a}\left(\theta_\mathrm{k}+\frac\pi2\right)\right|\psi\right\rangle\\&-\left\langle\psi\left|\mathrm{U}_\mathrm{a}^\dagger\left(\theta_\mathrm{k}-\frac\pi2\right)\mathrm{MU}_\mathrm{a}\left(\theta_\mathrm{k}-\frac\pi2\right)\right|\psi\right\rangle.
\end{aligned}
\end{equation}

This technique for computing gradients is called the parameter-shift rule. It enables the calculation of partial derivatives with respect to each parameter, which the optimizer then uses to update them.

The final loss combines the auxiliary classification losses from the first $K-1$ stages and the weighted final classification loss from the $K$-th stage. A supervision factor controls the contribution of auxiliary losses and adjusts the depth of supervision. The total loss function is given by equation \eqref{eq21}:

\begin{equation}
\label{eq21}
L=\frac{1}{\mathrm{K}\cdot\mathrm{T}}\left(\sigma\sum_{k=1}^{K-1}\sum_{t=1}^{T}L_{QACM}\left(p_{k}^{t},y\right)+\sum_{t=1}^{T} L_{\mathrm{SNN}}\left(p_{K}^{t},y\right)\right).
\end{equation}

During training, this study minimizes total loss and jointly optimizes the SNN and QNN parameters, thereby leveraging their respective strengths to improve performance.

\subsection{Overall algorithm steps and pseudocode}
The proposed QDS-SNN algorithm introduces a quantum-assisted classification layer for deep supervision and a new TSA-LIF neuron to enhance spatiotemporal modeling. The algorithm comprises four steps, as described in Algorithm 1 and illustrated in Fig. \ref{fig:1}.

\begin{algorithm}
\caption{The algorithm of QDS-SNN.}
\begin{algorithmic}
\STATE
\STATE \textbf{Input:} Traffic sign image $X\in R^{C\times H\times W}$, Time steps T, Stages K.
\STATE \textbf{Output:} Traffic sign classification result
\STATE  Direct coding: $X^{1:T} \stackrel{\mathrm{T}}{\leftarrow} X$ ;
\STATE \textbf{for} all T and K \textbf{do}
\STATE \hspace{0.5cm} Input $X^{1}$ into the Stage 1, passing through the Conv layer and TSA-LIF in turn;
\STATE \hspace{0.5cm} TSA-LIF receives: $H^{t-1,n}$, $X^{t,n}$;
\STATE \hspace{0.5cm} TSA-LIF outputs: \ensuremath{V^{t,n}{\leftarrow}\beta^n H^{t-1,n}+\alpha^n X^{t,n}}, 
\STATE where \ensuremath{\alpha^\mathrm{n}=\frac{\lambda^\mathrm{n}}{\tau_\mathrm{m}},\beta^\mathrm{n}=1-\frac{\mu^\mathrm{n}}{\tau_\mathrm{m}}};
\STATE \hspace{0.5cm} The outputs are fed into QAC layer and the next stage;
\STATE \hspace{0.5cm} Quantum encoding: 
$|$\ensuremath{\psi^{t}\rangle \overset{Amplitude Encoding}{\operatorname*{\leftarrow}}\ {x_i^t}};
\STATE \hspace{0.5cm} Perform quantum convolution and pooling;
\STATE \hspace{0.5cm} Measure the quantum state: \ensuremath{E_i(\theta) \leftarrow \langle\psi_k^t| Z|\psi_k^t\rangle};
\STATE \hspace{0.5cm} The classification results of the final stage are finally classified through GAP and FC layers.
\STATE \textbf{end for}
\STATE Compute QNN and SNN hybrid loss function: L= \\ 
\ensuremath{\frac{1}{\mathrm{K}\cdot\mathrm{T}}\left(\sigma\sum_{k=1}^{K-1}\sum_{t=1}^{T}L_{QACM}\left(\Theta\right)+\sum_{t=1}^{T} L_{\mathrm{CE}}\left(p_{K}^{t},y\right)\right)};
\STATE Update parameters\ensuremath{(\tau_\mathrm{m},\Theta,\sigma)}
\end{algorithmic}
\label{alg1}
\end{algorithm}

\textbf{Step 1.} Given the input $X\in R^{C\times H\times W}$ of the traffic sign recognition dataset, for the static input X in the dataset, first encode it directly to obtain $X \in R^{1:T}$ with a time dimension. This allows the static data to change dynamically over time, adapting to subsequent processing by the SNN.

\textbf{Step 2.} For $T$ time steps and $K$ stages, input $X^1$ into stage 1, processing it through the convolution layer and TSA-LIF. TSA-LIF integrates spatial features from the previous stage and temporal features from the prior time step, applying spatiotemporal attenuation factors to regulate feature retention and decay, thereby enhancing spatiotemporal modeling.

\textbf{Step 3.} TSA-LIF outputs temporal features through accumulation and excitation of membrane potential. These features enter the same stage at the next time step, while the feature map is sent to the QACM for auxiliary classification. QACM encodes inputs into quantum states and generates auxiliary classification results via quantum gate operations and measurement.

\textbf{Step 4.} In the final stage, the classification is performed through a global average pool and fully connected layers. The total loss incorporates a deep supervision factor to balance auxiliary losses. Parameter optimization is conducted using the proposed hybrid SNN-QNN training method.

\section{Experiment}
Here, the experimental results and performance of QDS-SNN on two representative traffic sign recognition datasets (GTSRB \cite{ref35} and TSRD \cite{ref36}) are presented. The specific model details are shown in Fig. \ref{fig:2}.

\subsection{Experimental setting}
\subsubsection{Datasets}
\begin{figure}
    \centering
    \includegraphics[width=0.8\linewidth]{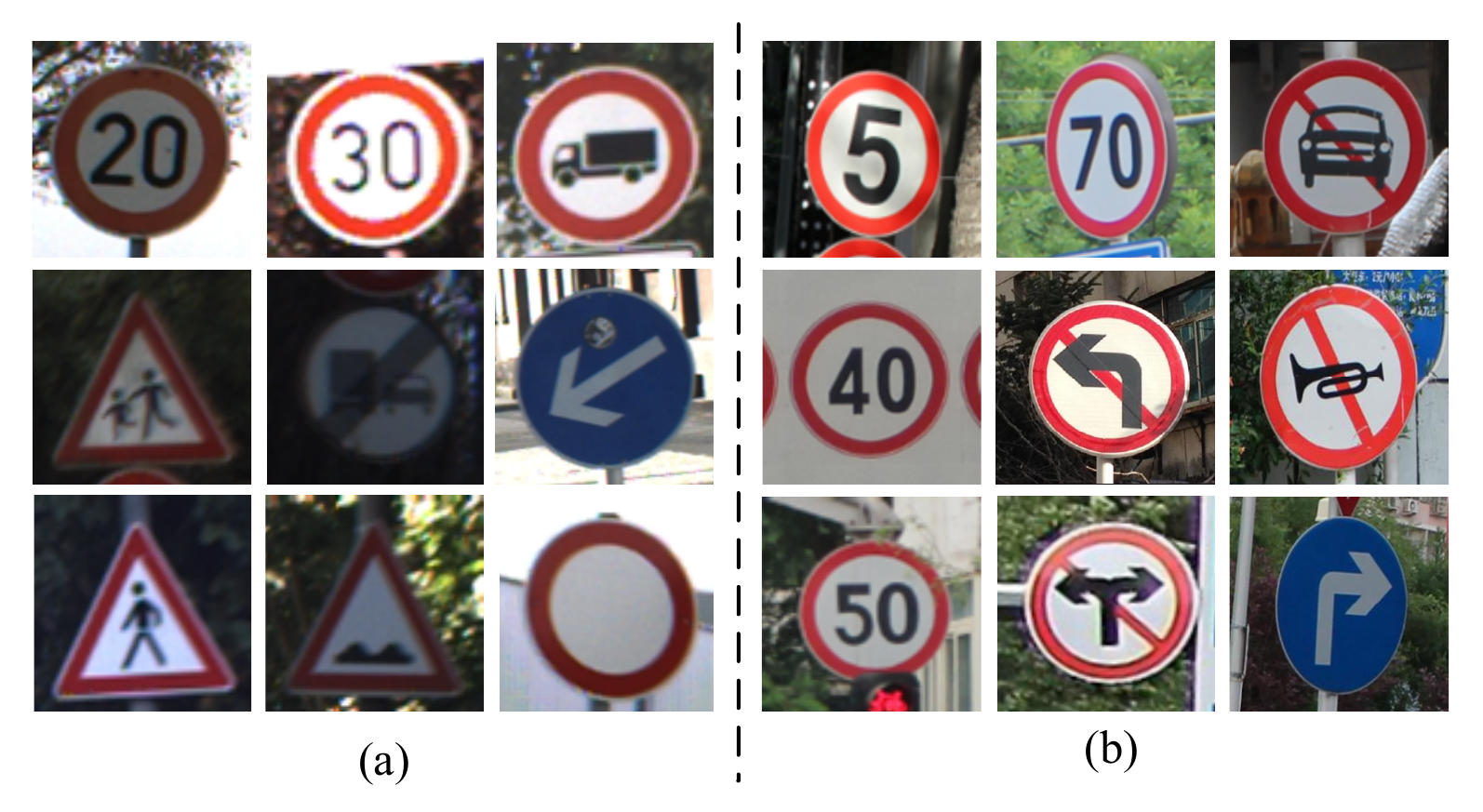}
    \caption{Traffic sign recognition dataset example. (a) GTSRB dataset example (b) TSRD dataset example}
    \label{fig:6}
\end{figure}

This experiment uses two popular datasets for traffic sign recognition: the German Traffic Sign Recognition Benchmark (GTSRB) and the Chinese Traffic Sign Database (TSRD). GTSRB contains over 50,000 images of 43 traffic sign types captured in real driving scenes, under varying angles and lighting conditions. TSRD comprises approximately 6,000 images of 58 common Chinese traffic signs under real-world conditions, including occlusions and blurring. These datasets enable a comprehensive evaluation of the model’s adaptability and accuracy across different traffic systems. Nine sample signs from each dataset are shown in Fig. \ref{fig:6}.

\subsubsection{Baselines}
To evaluate the effectiveness of the QDS-SNN model for traffic sign recognition, six representative SNN architectures are selected as baselines for comparison, as listed below:

1. ANN2SNN\cite{ref_ANN2SNN}: ANN2SNN maps pre-trained ANNs to SNNs for fast initialization but loses fine temporal details due to activation mismatch, reducing accuracy.

2. TET\cite{ref18}: TET improves SNN training by adjusting gradients to mitigate spiking errors, accelerating convergence.

3. tdBN\cite{ref17}: tdBN normalizes features across time and space using neuron thresholds, aligning with SNN dynamics to reduce distribution shifts and enhance the robustness to illumination and scale variations.

4. SEW-ResNet\cite{ref_SEW}: SEW-ResNet uses Spike-Element-Wise residual blocks with element-wise operations for identity mapping, allowing deeper networks and stable training.

5. SA-SCNN\cite{ref19}: SA-SCNN applies grayscale preprocessing and spatial attention to extract features, combining Poisson and neuron coding for improved performance.

6. Spikformer\cite{ref_Spikformer}: Spikformer integrates SNNs and Transformer with an efficient spiking self-attention mechanism, combining Transformer’s global modeling ability and SNNs’ low-energy advantage via event-driven computation.

7.MS-ResNet\cite{ref_MSResNet}: MS-ResNet adds residual connections on membrane potentials to enable direct information flow, mitigating gradient vanishing, and supporting multi-scale feature fusion, thus improving training stability in deep SNNs.

\subsubsection{Parameter setting}
The SNN training was performed on an NVIDIA RTX 3090 with 24 GB of memory, and the QNN training on an Intel Core i7 12400F. Table \ref{tab1} summarizes the key hyperparameters: 8 qubits were used for QNN encoding to balance representation capacity and mitigate barren plateau issues. The learning rate was set to 0.001, the batch size to 32, and training was carried out for 200 epochs. The Z measurement operator is used in the quantum circuit measurement. In the quantum simulation experiment, this study used PennyLane to construct and train quantum circuits, initializing them in the $|0\rangle$ state. The model's performance was evaluated on the GTSRB and TSRD datasets at time steps of 2, 4, and 6.
\begin{table}[htbp]
    \centering
    \caption{Hyperparameter settings in the experiment. (``T" means ``Time Steps"; ``DSF" means ``Deep Supervision Factor"; ``LR" means ``Learning Rate"; ``BS" means ``Batch Size")}
    \label{tab1}
    \resizebox{\columnwidth}{!}{ 
        \begin{tabular}{|c|c|c|c|c|c|c|}
            \hline
            Qubits & T & DSF $\sigma$ & LR & BS & Epoch & Measurement \\
            \hline
            8 & 2, 4, 6 & 0.8 & 0.001 & 32 & 200 & Z Gate \\
            \hline
        \end{tabular}
    }
\end{table}

\subsection{Experimental results}
\subsubsection{Experimental results and analysis of the GTSRB dataset} 
\begin{table*}
    \centering
    \caption{Comparison of the proposed method with previous methods working on GTSRB datasets.}
    \label{tab2}
        \begin{tabular}{|c|c|c|c|c|c|c|c|}
        \hline
        Methods & Architecture & Time Steps & Params(M) & Train Method & Power(mJ) & F1-Score(\%) & Accuracy(\%)\\
        \hline
        ANN2SNN\cite{ref_ANN2SNN} & ResNet-34 & 64 & 23.15 & ANN-SNN & - & - & 97.75\\
        \hline
        TET\cite{ref18} & Spiking-ResNet-34 & 6 & 23.15 & SG & - & 97.86 & 98.39 \\
        \hline
        tdBN\cite{ref17} & Spiking-ResNet-34 & 6 & 23.15 & SG & 6.39 & 94.96 & 95.45 \\
        \hline
        SEW-ResNet\cite{ref_SEW} & SEW-ResNet-34 & 4 & 23.15 & SG & 4.04 & 98.20 & 98.61 \\
        \hline
        SA-SCNN\cite{ref19} & SA-SCNN & 10 & - & SG & - & - & 99.56 \\
        \hline
        Spikformer\cite{ref_Spikformer} & Spikformer-8-768 & 4 & 68.71 & SG & 22.03 & 99.58 & 99.70 \\
        \hline
        \multirow{2}{*}{MS-ResNet\cite{ref_MSResNet}} & MS-ResNet-34 & 6 & 23.16 & SG & 5.11 & 98.22 & 98.40 \\
        \cline{2 - 8}
        & MS-ResNet-18 & 6 & 12.52 & SG & 4.29 & 96.40 & 96.86 \\
        \hline
        \multirow{3}{*}{QDS-SNN} & \multirow{3}{*}{QDS-SNN} & 6 & 17.76 & SG & 2.26 & \textbf{99.67} & \textbf{99.72} \\
        \cline{3 - 8}
        & & 4 & 17.76 & SG & 1.46 & 99.12 & 99.43 \\
        \cline{3 - 8}
        & & 2 & 17.76 & SG & 1.03 & 98.49 &98.68 \\
        \hline
    \end{tabular}
\end{table*}

As shown in Table \ref{tab2}, QDS-SNN shows a significant performance improvement compared to other SNN-based traffic sign recognition models. The table compares the performance of methods, including ANN2SNN, TET, tdBN, SEW-ResNet, SA-SCNN, Spikformer, and MS-ResNet, on the GTSRB dataset. The results show that QDS-SNN achieves superior performance in accuracy, F1-score, energy consumption, and time step.

On the GTSRB dataset, QDS-SNN achieved 98.82\% accuracy using only 2 time steps, outperforming most existing methods. Consistent with the low-latency advantage afforded by fewer time steps in \cite{ref16}, this result shows that the model achieves excellent performance at low time steps. Its core advantage lies in achieving high performance and low energy consumption through collaborative optimization in resource-constrained intelligent transportation scenarios. At 6 time steps, QDS-SNN achieved a maximum accuracy of 99.72\%, outperforming MS-ResNet (T=6) by 1.32\% and SEW-ResNet by 1.11\%. Even compared to SA-SCNN, which incorporates spatial attention, QDS-SNN maintains an accuracy advantage. At the same time step (T=4), the proposed QDS-SNN model achieved an accuracy of 99.43\%, while Spikformer achieved 99.70\%. QDS-SNN is slightly lower than Spikformer's, but it still maintains a high recognition accuracy.

In terms of energy consumption, QDS-SNN also shows excellent efficiency. While maintaining high accuracy, it has significantly lower power consumption than other models, at only 2.26 mJ over 6 time steps, compared with 6.39 mJ for tdBN and 5.11 mJ for MS-ResNet. When the time step is reduced to 4 and 2, the power consumption decreases to 1.46 and 1.03 mJ, respectively, demonstrating a significant energy advantage. Although Spikformer performs slightly better than QDS-SNN at the same time step (T=4), QDS-SNN has significant advantages in energy consumption and parameter count.

In summary, QDS-SNN not only achieves higher recognition accuracy and F1-score on the GTSRB dataset but also significantly reduces power consumption. It also offers significant advantages in time-step control and model compression, demonstrating strong deployment potential and practical applicability.

\subsubsection{TSRD dataset experimental results and analysis} 

\begin{table*}
    \centering
    \caption{Comparison of the proposed method with previous methods working on TSRD datasets.}
    \label{tab3}
        \begin{tabular}{|c|c|c|c|c|c|c|c|}
        \hline
        Methods & Architecture & Time Steps & Params(M) & Train Method & Power(mJ) & F1-Score(\%) & Accuracy(\%)\\
        \hline
        ANN2SNN\cite{ref_ANN2SNN} & ResNet-34 & 64 & 23.16 & ANN-SNN & - & - & 96.12\\
        \hline
        TET\cite{ref18} & Spiking-ResNet-34 & 6 & 23.16 & SG & - & 96.85 & 97.31 \\
        \hline
        tdBN\cite{ref17} & Spiking-ResNet-34 & 6 & 23.16 & SG & 6.63 & 94.22 & 95.18 \\
        \hline
        SEW-ResNet\cite{ref_SEW} & SEW-ResNet-34 & 4 & 23.16 & SG & 4.32 & 96.58 & 97.06 \\
        \hline
        SA-SCNN\cite{ref19} & SA-SCNN & 10 & - & SG & - & - & 97.16 \\
        \hline
        Spikformer\cite{ref_Spikformer} & Spikformer-8-768 & 4 & 68.73 & SG & 23.34 & 97.52 & 97.85 \\
        \hline
        \multirow{2}{*}{MS-ResNet\cite{ref_MSResNet}} & MS-ResNet-34 & 6 & 23.17 & SG & 5.32 & 96.51 & 97.03 \\
        \cline{2 - 8}
        & MS-ResNet-18 & 6 & 12.53 & SG & 4.41 & 95.20 & 95.80 \\
        \hline
        \multirow{3}{*}{QDS-SNN} & \multirow{3}{*}{QDS-SNN} & 6 & 17.77 & SG & 2.38  & \textbf{97.68} & \textbf{97.90} \\
        \cline{3 - 8}
        & & 4 & 17.77 & SG & 1.67 & 96.99 & 97.21 \\
        \cline{3 - 8}
        & & 2 & 17.77 & SG & 1.12 & 96.50 & 96.89 \\
        \hline
    \end{tabular}
\end{table*}
As shown in Table \ref{tab3}, QDS-SNN also performs well in the TSRD dataset, outperforming state-of-the-art methods. This table compares key indicators, including accuracy, F1 Score, number of parameters, energy consumption, and time step, for several representative SNN models on the TSRD dataset. Compared to ANN2SNN, TET, tdBN, SEW-ResNet, SA-SCNN, and MS-ResNet, QDS-SNN significantly reduces time delay and energy consumption while maintaining high accuracy.

On the TSRD dataset, QDS-SNN achieves 96.89\% classification accuracy with only 2 time steps, exceeding those of ANN2SNN, tdBN, and MS-ResNet-18. When the time step is 4, the accuracy of QDS-SNN is further improved to 97.21\%, an increase of 2.03\% over tdBN and 0.15\% over SEW-ResNet. At the same time, energy consumption is only 1.67 mJ, significantly lower than 6.63 mJ for tdBN and 5.32 mJ for MS-ResNet-34. At the same time step (T=4), the proposed QDS-SNN model achieved an accuracy of 97.21\%, while Spikformer achieved 97.85\%. QDS-SNN is slightly lower than Spikformer's but still maintains high recognition accuracy.

It should be noted that at 6 time steps, QDS-SNN achieved a maximum accuracy of 97.90\%, surpassing the current MS-ResNet-34 and MS-ResNet-18 layers, with improvements of 0.87\% and 2.10\%, respectively, while maintaining extremely low parameter count (only 17.77M) and energy consumption (2.38 mJ), fully demonstrating the comprehensive advantages of this method in accuracy, efficiency, and model scale. Although Spikformer performs slightly better than QDS-SNN at the same time step (T=4), QDS-SNN has significant advantages in energy consumption and parameter count.

In summary, QDS-SNN maintains high accuracy and F1-Score even with a low number of time steps on the TSRD dataset, while significantly reducing energy consumption, demonstrating strong practicality and value for edge deployment in real-world applications.

\subsubsection{Ablation experiment}
\paragraph{Effects of spatiotemporal decay factors on neurons}
\begin{table}
    \centering
    \caption{Ablation study of neuron spatiotemporal decay factors on GTSRB dataset.}
    \label{tab4}
        \begin{tabular}{|c|c|c|c|c|}
        \hline
        Neuron & T & $\alpha^n$ & $\beta^n$ & Acc.(\%) \\
        \hline
        Normal LIF neuron & 6 & $\times$ & $\times$ & 98.85 \\
        \hline
        with $\alpha^n$ only & 6 & $\checkmark$ & $\times$ & 99.56 \\
        \hline
        with $\beta^n$ only & 6 & $\times$ & $\checkmark$ & 99.20 \\
        \hline
        TSA-LIF neuron & \multirow{2}{*}{6} & \multirow{2}{*}{$\checkmark$} & \multirow{2}{*}{$\checkmark$} & \multirow{2}{*}{\textbf{99.72}} \\
        with both $\alpha^n$ and $\beta^n$ & &  & & \\
        \hline
    \end{tabular}
\end{table}

To analyze the impact of the spatial decay factor $\alpha^n$ and the temporal decay factor $\beta^n$ on model performance, ablation experiments were conducted on the GTSRB dataset with T=6 (Table \ref{tab4}). The evaluation compared standard LIF neurons (no decay), neurons with only spatial decay, only temporal decay, and TSA-LIF neurons with both spatial and temporal decay. The standard LIF achieved 98.85\% accuracy; adding spatial decay improved it to 99.56\%, highlighting the benefits of enhanced feature selection; adding temporal decay reached 99.20\%, showing the benefits of temporal modeling.
Combining both decay factors in TSA-LIF yielded the highest accuracy of 99.72\%, surpassing all single-factor variants and demonstrating the synergistic effect of modeling spatial and temporal dependencies. These results confirm TSA-LIF’s effectiveness in improving neuronal expressiveness and recognition accuracy while maintaining a low number of time steps, providing a reference for building high-performance SNNs under energy constraints. 

\paragraph{Effects of deep supervision}
\begin{table}
    \centering
    \caption{Ablation study of deep supervision on GTSRB dataset.}
    \label{tab_DeepSupervision}
        \begin{tabular}{|c|c|c|c|}
        \hline
         & \multirow{2}{*}{T} & \multirow{2}{*}{Deep Supervision} & Acc. \\
         & & & (\%) \\
        \hline
        QDS-SNN & \multirow{2}{*}{6} & \multirow{2}{*}{$\times$} & \multirow{2}{*}{98.80} \\
         (without deep supervision) & & & \\
        \hline
        QDS-SNN & \multirow{2}{*}{6} & \multirow{2}{*}{Classic Deep Supervision} & \multirow{2}{*}{99.70} \\
        (with CACM) & & & \\
        \hline
        QDS-SNN & \multirow{2}{*}{6} & \multirow{2}{*}{Quantum Deep Supervision} & \multirow{2}{*}{\textbf{99.72}} \\
         (with QACM) & & & \\
        \hline
    \end{tabular}
\end{table}
To verify the effectiveness of quantum deep supervision, this study conducted ablation experiments on the GTSRB dataset (Table \ref{tab_DeepSupervision}). Ablation experiments were performed in three scenarios: without deep supervision, with a quantum-assisted classification module (QACM), and with a classical auxiliary classification module (CACM). QACM uses quantum convolution and pooling for auxiliary classification and leverages quantum computing's natural low energy consumption to optimize deep supervision. CACM consists of three Conv layers, followed by a GAP layer and an FC layer to produce traffic sign classification results. The model using QACM achieved the highest accuracy of 99.72\%, while the model using CACM achieved a similar performance of 99.70\%, although its energy consumption was significantly higher than QACM's. Furthermore, removing the deep supervision module reduced accuracy to 98.80\%.

These results demonstrate that quantum deep supervision can alleviate gradient vanishing and information loss, enhance learning of intermediate features and multi-scale fusion, and improve recognition accuracy. The performance improvement confirms the effectiveness of staged deep supervision in optimizing SNN training, highlighting its importance for training efficiency and final performance.

\begin{figure}
    \centering
    \includegraphics[width=0.7\linewidth]{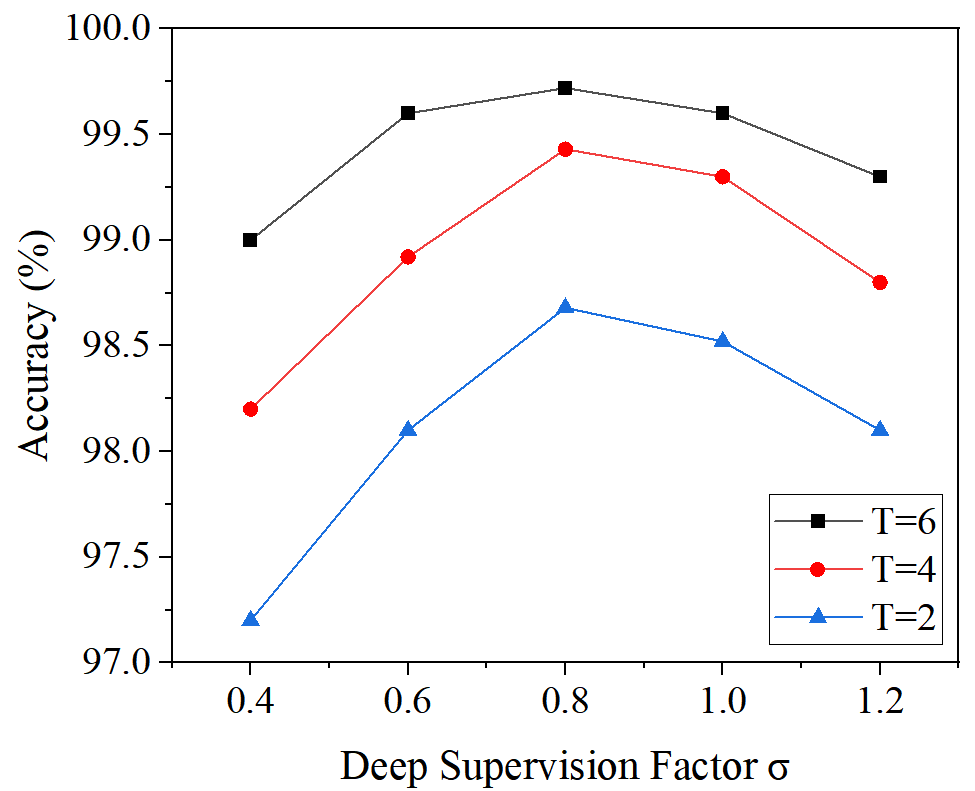}
    \caption{Ablation study of deep supervision factor $\sigma$ on GTSRB dataset.}
    \label{fig:7}
\end{figure}
To explore the impact of the deep supervision factor $\sigma$ on model performance, an ablation study is conducted on the GTSRB dataset, and the results are shown in Fig. \ref{fig:7}. The model inserts a quantum-assisted classification layer (QAC) after each stage to implement deep supervision. The deep supervision loss function is defined in equation \eqref{eq20}, where $\sigma$ controls the weight of auxiliary losses from earlier stages in the total loss.

As $\sigma$ increases from 0.4, model performance improves, reaching a peak at $\sigma=0.8$. Beyond this point, the accuracy slowly declines. This indicates that $\sigma=0.8$ best balances the auxiliary and final loss. Deep supervision guides feature learning through auxiliary classifiers; however, if $\sigma$ is too large, early-stage features, which are still undertrained and unreliable, over-influence the loss computation, leading the learning process towards local features and neglecting global fusion, thereby reducing robustness. In contrast, a too-small $\sigma$ weakens auxiliary supervision, limits feature learning, and hinders accuracy improvement. Therefore, $\sigma=0.8$ is set as the optimal value in this study.

\subsection{Performance Analysis}
\subsubsection{Energy consumption analysis}
FLOPs are used to estimate a CNN's computational energy consumption, whereas SNN energy calculation is more nuanced. In our QDS-SNN, only the input encoding and final FC classification layers run full floating-point operations; all other Conv and spiking layers adopt event-driven computation, dominated by AC operations triggered by spike propagation. The QDS-SNN architecture converts dense matrix multiplications into sparse additions. The specific computational consumption metrics for processing a single sample from the GTSRB and TSRD datasets (T=6) are detailed in Table \ref{tab_Computation}, including key indicators such as total spike count, AC, MACs, FLOPs, and SOPs that directly inform the energy calculation. For QNN in the QACM module, the transformation of quantum unitary gates is a coherent reversible operation with negligible energy consumption, a widely accepted simplification in quantum computing energy modeling \cite{ref26}. Specifically, the energy consumption of unitary gate operations is much lower than that of quantum state preparation, measurement, and other irreversible processes, and its contribution to total energy consumption can be safely ignored \cite{ref38}. Thus, this study neglects the energy consumption of quantum unitary gate operations in our energy model, focusing instead on the classical SNN operations that dominate the total inference energy. The inference energy consumption of QDS-SNN can be expressed as equation (18):

\begin{equation}
\label{eq20}
\begin{aligned}
E_{total}=&E_{QACM}+E_{MAC}\cdot FL_{en} \\& +E_{AC}\cdot T\cdot(\sum_{m=1}^MFL_{Com\nu}^m\cdot fr^m \\& +\sum_{n=1}^NFL_{GAP}^n\cdot fr^n+FL_{FC}^K\cdot fr^K)
\end{aligned}
\end{equation}

\begin{table}
    \centering
    \caption{Computational consumption for processing a single sample of GTSRB and TSRD.}
    \label{tab_Computation}
        \begin{tabular}{|c|c|c|c|c|}
        \hline
        Dataset & GTSRB & TSRD  \\
        \hline
        Time Steps & 6 & 6 \\
        \hline
        ACs($\times 10^9$) & 2.48 & 2.62 \\
        \hline
        MACs($\times 10^6$) & 38.0 & 40.4 \\
        \hline
        FLOPs($\times 10^9$) & 2.56 & 2.70 \\
        \hline
        SOPs($\times 10^9$) & 2.42 & 2.56 \\
        \hline
        Spikes($\times 10^6$) & 11.8 & 12.5 \\
        \hline
        Params(M) & 17.76 & 17.77 \\
        \hline
        Energy(mJ) & 2.26 & 2.38 \\
        \hline
    \end{tabular}
\end{table}

Among them, $M$ and $N$ denote the total number of Conv and GAP layers, respectively. $E_{MAC}$ and $E_{AC}$ represent the energy cost of MAC and AC operations. $fr^m$, $fr^n$, $FL^n_{Conv}$, and $FL^m_{GAP}$ represent the firing rate and FLOPs of the $n$-th Conv layer and the $m$-th GAP layer, respectively. $FL_{en}$ denotes the FLOPs of the encoding layer, while $fr^K$ and $FL^{K}_{FC}$ denote the firing rate and the FLOPs of the $K$-th FC layer. The previous SNN study\cite{ref37} assumed 45nm CMOS technology for 32-bit floating-point implementation, where $E_{MAC}=4.6$ pJ and $E_{AC}=0.9$ pJ for various operations.

\begin{table}
    \centering
    \caption{Comparison of performance and energy consumption of CACM and QACM on the GTSRB dataset.}
    \label{tab6}
    \resizebox{\columnwidth}{!}{ 
        \begin{tabular}{|c|c|c|c|c|c|}
        \hline
        Module & Method & T & Params(M) & Power(mJ) & Acc.(\%) \\
        \hline
        CACM & Conv+FC & 6 & 24.60 & 6.73 & 99.70 \\
        \hline
        QACM & QACM & 6 & \textbf{17.76} &  \textbf{2.26}  & \textbf{99.72} \\
        \hline
    \end{tabular}
    }
\end{table}
As shown in Table \ref{tab6}, the energy consumption of the quantum-assisted classifier (QACM) constructed by quantum convolution, pooling, etc., compared to the classical-assisted classifier (CACM) constructed by traditional ``Conv+FC", verifies the advantages of the quantum module. At the same time (T=6), the parameters, energy consumption, and classification accuracy are compared.

The experimental results, shown in Table \ref{tab6}, demonstrate that both QACM and CACM achieved highly comparable accuracies in the GTSRB dataset (QACM: 99.72\%, CACM: 99.70\%), with QACM showing a slight advantage of 0.02\% over CACM. However, despite the small difference in accuracy, QACM significantly reduced the number of parameters by 27.8\% and energy consumption by 66.4\%. These results validate the core value of QACM: it achieves low energy consumption and low parameter optimization without sacrificing accuracy, exhibiting significant advantages in both model size and energy efficiency. For real-time edge tasks such as traffic sign recognition, QACM not only alleviates computational and power-consumption limitations but also retains the functionality of deep supervision, making it a promising direction for lightweight and energy-efficient traffic perception systems.

\subsubsection{Convergence analysis}
\begin{figure}
    \centering
    \includegraphics[width=0.8\linewidth]{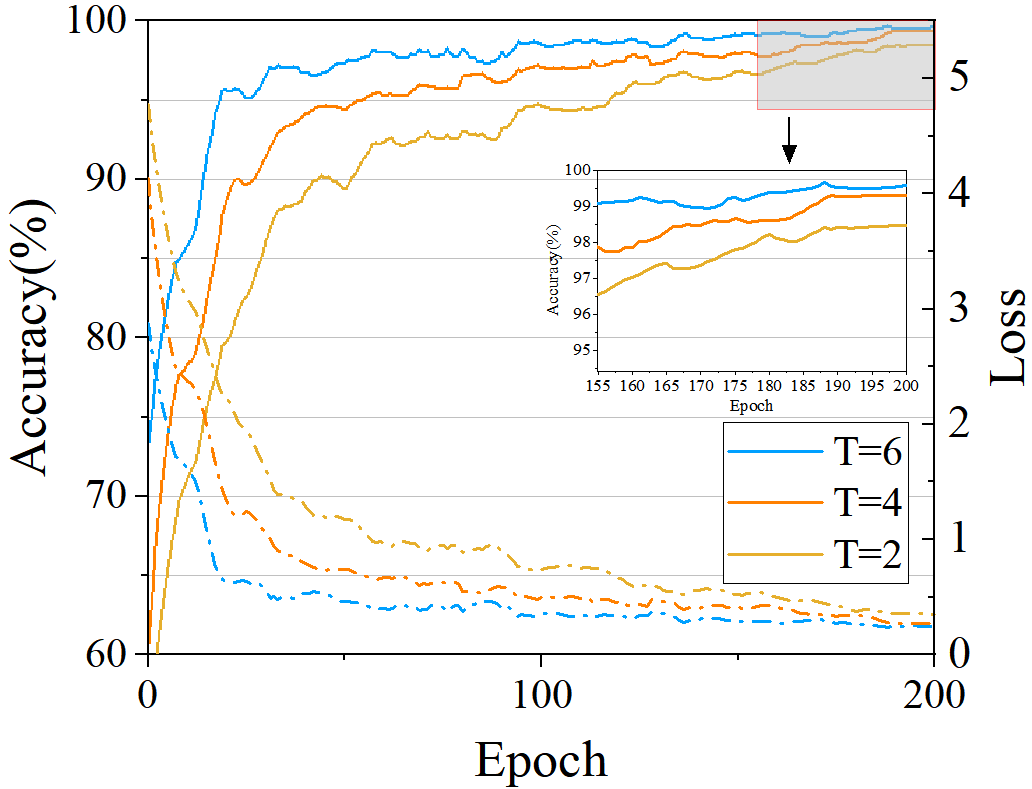}
    \caption{Convergence comparison of QDS-SNN on different time steps on the GTSRB dataset.}
    \label{fig:8}
\end{figure}
\begin{figure}
    \centering
    \includegraphics[width=0.8\linewidth]{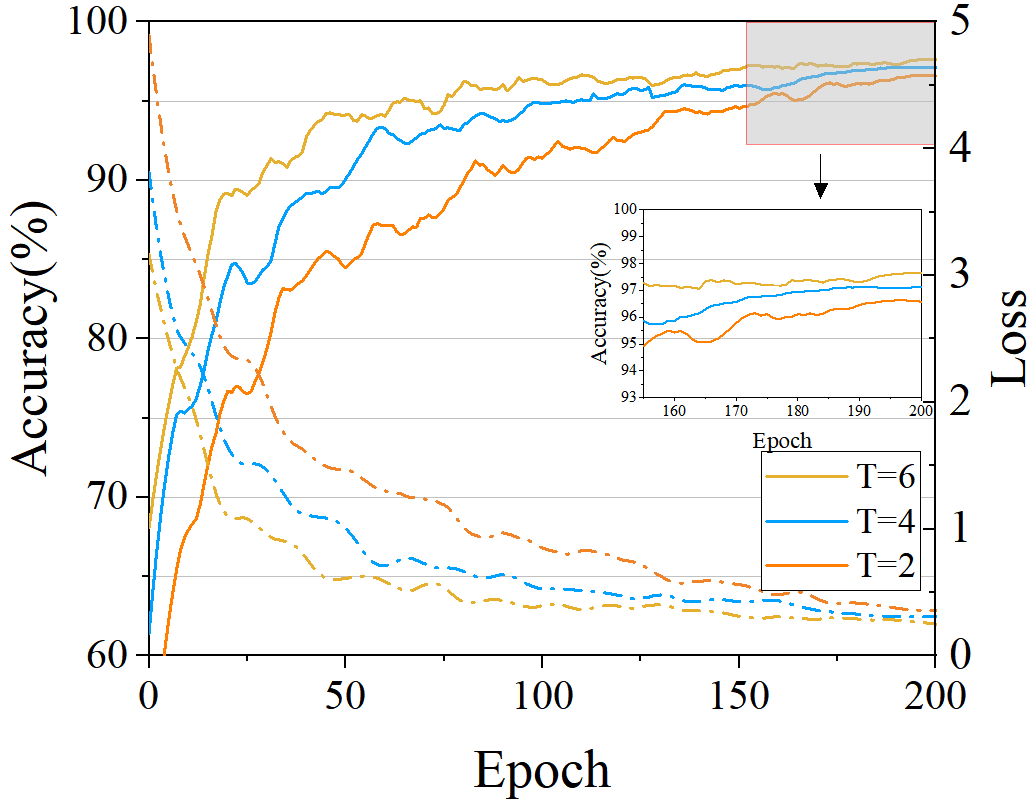}
    \caption{Convergence comparison of QDS-SNN on different time steps on the TSRD dataset.}
    \label{fig:9}
\end{figure}
The convergence of our proposed QDS-SNN is demonstrated in Figs. \ref{fig:8} and \ref{fig:9}. The model's convergence is compared at different time steps (T=2, T=4, T=6) on the GTSRB (Fig. \ref{fig:8}) and TSRD (Fig. \ref{fig:9}) datasets. As can be seen from the two figures, when T=2, T=4, and T=6, as the training rounds (epochs) increase, the model accuracy (Accuracy) shows a steady upward trend, which fully demonstrates that QDS-SNN can effectively carry out the learning process on the two types of traffic sign datasets, GTSRB and TSRD, and the advancement of training depth can continuously improve the model performance.

The convergence curves on the GTSRB dataset reveal that longer simulation time steps improve training efficiency and accuracy. At T=6, the model rapidly surpasses 95\% accuracy in 50 epochs and approaches optimal performance before 100 epochs. With T=4, the convergence slows slightly, but the accuracy remains high. At T=2, the convergence is much slower, taking nearly 200 epochs to stabilize, and the final precision is lower.

Similar trends are observed on the TSRD dataset. Longer time steps facilitate richer spike propagation and temporal integration, enhancing early convergence and feature learning. In contrast, shorter time steps restrict temporal dynamics, leading to slower learning and reduced performance. In general, increasing the time step improves spatiotemporal feature extraction and model stability, providing valuable guidance for configuring QDS-SNN in practical applications.

In summary, QDS-SNN shows faster convergence and higher performance on the GTSRB and TSRD datasets with T=6 and T=4; the convergence rate slows with T=2 but eventually achieves higher performance after more training epochs. This shows the effectiveness of the proposed QDS-SNN algorithm.

\subsubsection{Complexity analysis}
The complexity of the proposed QDS-SNN is analyzed and primarily comprises SNN complexity and QAC layer complexity. The SNN complexity includes $T$ time steps and $K$ stages. For a single time step, the complexity of the convolutional layer and the TSA-LIF neuron must be calculated separately. Assuming that the input feature map size is $H\times W,$ the number of channels is $C_{in}$, the convolution kernel size is $K_{size}\times K_{size}$, and the number of output channels is $C_{out}$, then the complexity of a single convolution layer is $\mathrm{O(K_{size}^2\cdot C_{in}^k\cdot C_{out}^k\cdot H\cdot W)}$. TSA-LIF neurons include membrane potential updates and pulse excitations, which are essentially linear transformations by elements and threshold judgments, with a complexity of $\mathrm{O(C_{out}^{k}\cdot H\cdot W)}$. Therefore, the total complexity of a single time step and a single stage is $\mathrm{O(K_{size}^{2}\cdot C_{in}^{k}\cdot C_{out}^{k}\cdot H\cdot W+C_{out}^{k}\cdot H\cdot W)}$. For $T$ time steps and $K$ stages, the total complexity must be multiplied by the number of stages K and the time step T: $\mathrm{O(T\cdot K\cdot K_{size}^{2}\cdot C_{in}^{k}\cdot C_{out}^{k}\cdot H\cdot W+T\cdot K\cdot C_{out}^{k}\cdot H\cdot W)}$, that is, $O(T\cdot\mathrm{K}\cdot\mathrm{H}\cdot\mathrm{W}\cdot\mathrm{C_{in}\cdot C_{out}})$.

The complexity of the QAC layer encompasses both the quantum circuit complexity and the time complexity. The complexity of quantum circuits is typically measured by the number of quantum gates involved. As illustrated in Figs. \ref{fig:4} and \ref {fig:5}, the 4-qubit quantum convolution and pooling layers require 10 and 8 quantum gates, respectively, while the 2-qubit versions use 5 and 4 gates. Given that QDS-SNN consists of $K$ stages, with each of the first $K-1$ stages followed by a QAC layer, the total number of quantum gates can be computed: $\mathrm{(K-1)\cdot T\cdot(10\cdot2+8\cdot2+5\cdot3+4\cdot2)=59(K-1)\cdot T}$.

Thus, the overall computational complexity of all QNN modules is on the order of $O(K \cdot T)$. Lower time complexity typically reflects greater computational efficiency, thus reducing network resource demands. When encoding classical data of dimension $n$ into quantum states, the standard complexity is $O(n)$. However, by leveraging amplitude encoding, QNNs can represent an $n$-dimensional vector using only $\lceil log_{2}n \rceil$ qubits. This allows the model's time complexity to be reduced to $O(log_{2}n)$, highlighting the efficiency gains brought about by the quantum representation.

In summary, the overall complexity of the proposed QDS-SNN algorithm is $\mathrm{O(T\cdot K\cdot H\cdot W\cdot C_{in}\cdot C_{out})}$.

\subsubsection{Quantum Noise Fidelity Analysis}
In the NISQ era, QML models will face quantum noise, which will reduce accuracy and prevent convergence. Single-qubit noise is described by a Kraus matrix, which includes four types of noise: Bit Flip (BF), Phase Flip (PF), Amplitude Damping (AD), and Depolarizing Noise (DN) \cite{ref39}.

\begin{figure}
    \centering
    \includegraphics[width=0.49\linewidth]{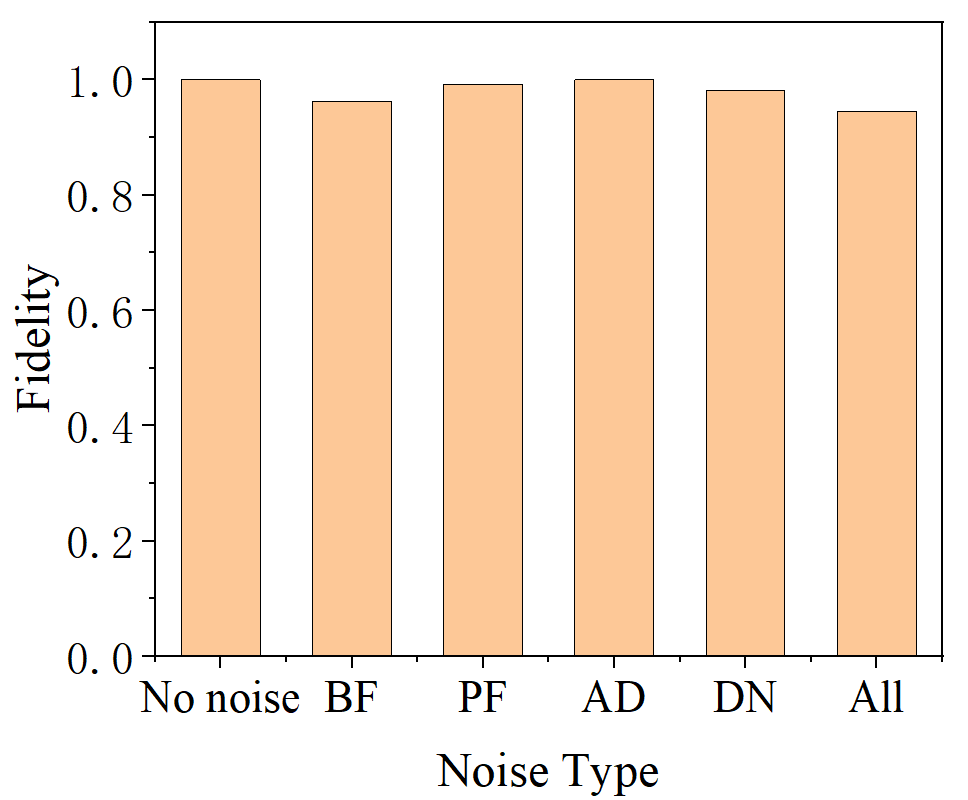}
    \includegraphics[width=0.49\linewidth]{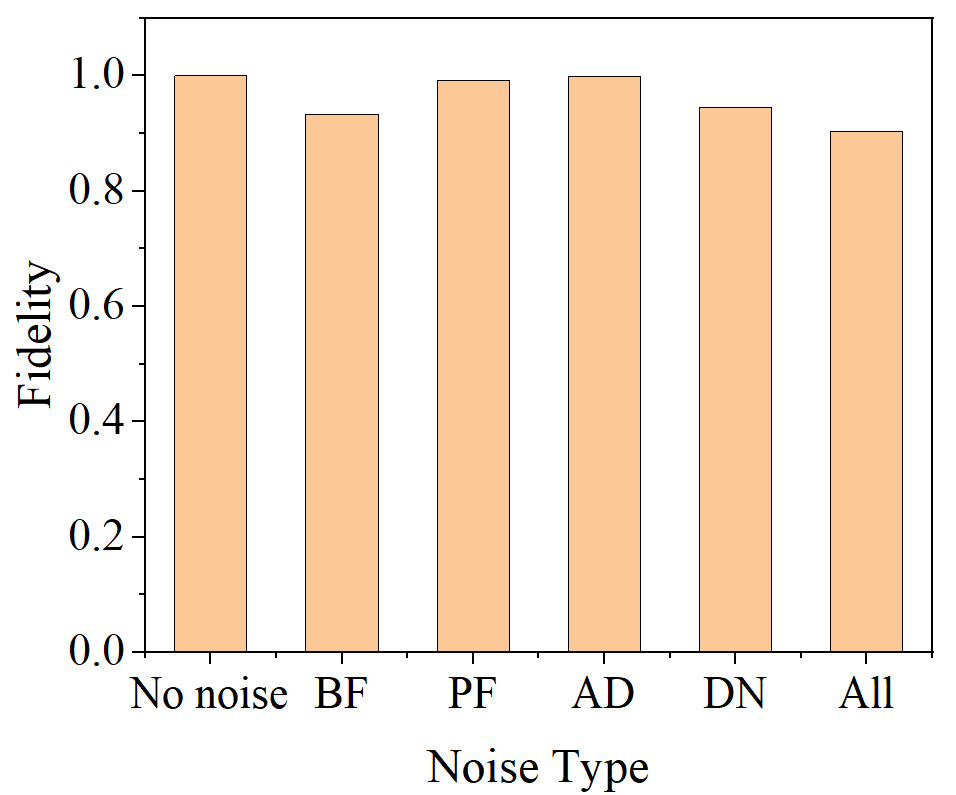}
    \caption{Fidelity of QNN under different quantum noise when p=0.01(left) and p=0.1(right).}
    \label{fig_noise}
\end{figure}

Fidelity measures the robustness of a quantum algorithm to noise by quantifying the similarity between two quantum states. It reflects the probability that one state will be recognized as another upon measurement. A high-fidelity quantum algorithm can better tolerate noise, indicating stronger reliability and stability. The formula for estimating fidelity between two mixed states is given by equation \eqref{eq24}.
\begin{equation}
\begin{aligned}
\label{eq24}
\mathrm{F}(\rho,\sigma)=\mathrm{Tr}(\sqrt{\sqrt{\rho}\sigma\sqrt{\rho}})^2
\end{aligned}
\end{equation}
Here, $\rho$ and $\sigma$ represent the density matrices of the two mixed states. In this work, the fidelity of the QNN circuit was evaluated under four individual noise channels and a composite noise channel at noise probabilities p=0.01 and p=0.1. ``All" denotes a composite noise channel, which applies a superposition of all single-qubit noise operations (Kraus matrices). The corresponding results are summarized in Table \ref{tab_Noise} and illustrated in Fig. \ref{fig_noise}.
\begin{table}
    \centering
    \caption{Fidelity of QNN under different quantum noises when p=0.01 and p=0.1.}
    \label{tab_Noise}
        \begin{tabular}{|c|c|c|c|c|}
        \hline
        Noise type & p=0.01 & p=0.1\\
        \hline
        No noise &   1       &   1           \\
        \hline
        BF      &   0.9625  &   0.9330    \\
        \hline
        PF      &   0.9920  &   0.9915     \\
        \hline
        AD      &   0.9996  &   0.9991    \\
        \hline
        DN      &   0.9815  &   0.9453      \\
        \hline
        All     &   0.9452  &   0.9037   \\
        \hline
    \end{tabular}
\end{table}

\section{Conclusion}
This paper proposes QDS-SNN, an energy-efficient traffic sign recognition algorithm based on quantum deep supervised SNNs. Traditional CNN-based methods require high computational and energy resources, limiting their use in edge devices. QDS-SNN integrates temporally and spatially adaptive spiking neurons (TSA-LIF) with quantum-assisted classification modules (QACM), thereby introducing quantum deep supervision to mitigate vanishing gradients and improve training efficiency. Experiments show that QDS-SNN achieves 99.72\% accuracy on GTSRB with only 6 time steps—1.32\% higher than the MS-ResNet baseline—while reducing energy consumption by 55.77\%. In TSRD, it achieves 97.90\% accuracy while consuming 52.68\% of the baseline energy. By combining quantum computing with SNN, QDS-SNN achieves high accuracy and low power consumption, making it suitable for real-time intelligent traffic perception on resource-constrained platforms.

The current QDS-SNN model has room for improvement. Future work will extend from image classification to multi-scale traffic target detection by more deeply integrating quantum attention mechanisms. This approach retains the advantages of QDS-SNN for efficient and low-energy recognition while enabling real-time detection in complex scenes (e.g., mixed speed limits and warning signs). Additionally, the algorithm will be adapted for multimodal traffic scenarios by integrating visual, radar, and other data to improve generalization in complex environments, advancing QSNN development in intelligent transportation systems.






\vspace{+0.5 cm}
{\bf Zhiguo Qu} is currently a Professor in the School of Computer Science. His research interests include quantum machine learning, quantum communications, and quantum computing.

\vspace{+0 cm}
{\bf Keqi Li} is currently a master student at Nanjing University of Information Science and Technology. His research interests include Internet of Vehicles and quantum machine learning.

\vspace{+0 cm}
{\bf Le Sun} is currently a Professor in the School of Computer and Software, Nanjing University of Information Science and Technology, China. Her research interests include medical data stream mining, services
computing, and fog computing.

\vspace{+0 cm}
{\bf Wenjie Liu} is a Professor of Computer Science at Nanjing University of Information Science and Technology. His research interests include quantum machine learning, quantum secure multiparty computation, and quantum communication.

\vspace{+0 cm}
{\bf Yimin Yu}, Ph.D., is a Senior Engineer and serves as the Dean of the School of Information at Yunnan University of Finance and Economics, the Director of the Yunnan Provincial International Joint Research and Development Center for Cross-Border Trade and Finance, and the President of the Yunnan Provincial Electronic Governance Association.

\vspace{+0 cm}
{\bf Saif Al-Kuwari} is currently an Associate Professor in the College of Science and Engineering at Hamad Bin Khalifa University, Qatar. His research interests include quantum computing, cryptography, and computational forensics.

\vspace{+0 cm}
{\bf Ahmed Farouk} is currently an Associate Professor at Hurghada University, Egypt. His research interests include theories of quantum information, machine learning, and cryptography.


 






\begin{thebibliography}{1}
\bibliographystyle{IEEEtran}

\bibitem{ref1}
A. Shustanov and P. Yakimov, "CNN design for real-time traffic sign recognition," \textit{Procedia Eng.}, vol. 201, pp. 718–725, 2017.

\bibitem{ref2}
Z. Qu, Y. Xia, L. Sun, W. Liu, and G. Muhammad, "QCACNN: A quantum convolutional neural network algorithm for traffic sign recognition in carbon-intelligent electric vehicles," \textit{IEEE Internet Things J.}, vol. 12, no. 17, pp. 34708–34719, 2025.


\bibitem{ref_ITSRS}
M. Q. Kheder and A. A. Mohammed, "Improved traffic sign recognition system (ITSRS) for autonomous vehicle based on deep convolutional neural network," \textit{Multimed. Tools Appl.}, vol. 83, no. 22, pp. 61821–61841, 2024.

\bibitem{ref6}
A. Shustanov and P. Yakimov, "CNN design for real-time traffic sign recognition," \textit{Procedia Eng.}, vol. 201, pp. 718–725, 2017.

\bibitem{ref7}
J. Hu, L. Shen, and G. Sun, "Squeeze-and-excitation networks," in \textit{Proc. IEEE Conf. Comput. Vis. Pattern Recognit.}, pp. 7132–7141, 2018.

\bibitem{ref8}
N. Hasan, T. Anzum, and N. Jahan, "Traffic sign recognition system (TSRS): SVM and convolutional neural network," in \textit{Invent. Commun. Comput. Technol.: Proc. ICICCT 2020}, pp. 69–79, 2020.

\bibitem{ref9}
Z. Lian, Q. Zeng, W. Wang, D. Xu, W. Meng, and C. Su, “Traffic sign recognition using optimized federated learning in internet of vehicles,” \textit{IEEE Internet Things J.}, vol. 11, no. 4, pp. 6722–6729, 2023.

\bibitem{ref10}
L. You, Z. Guo, B. Zuo, Y. Chang, and C. Yuen, “SLMFed: A stage‑based and layerwise mechanism for incremental federated learning to assist dynamic and ubiquitous IoT,” \textit{IEEE Internet Things J.}, vol. 11, no. 9, pp. 16364–16381, 2024.

\bibitem{ref10_1}
K. Xie, Z. Zhang, Bo. Li, et al., "Efficient federated learning with spike neural networks for traffic sign recognition," \textit{IEEE Trans. Veh. Technol.}, vol. 71, no. 9, pp. 9980–9992, 2022.

\bibitem{ref10_4}
Y. Zhang, H. Xu, L. Huang, and C. Chen, "A storage-efficient SNN--CNN hybrid network with RRAM-implemented weights for traffic signs recognition," \textit{Eng. Appl. Artif. Intell.}, vol. 123, p. 106232, 2023.

\bibitem{ref10_2}
H. Chen, Y. Liu, W. Ye, and C. Han, “Traffic sign recognition model based on spiking neural network,” in \textit{2024 4th Int. Conf. Neural Netw., Inf. Commun. Eng. (NNICE 2024)}, pp. 693–696, 2024.

\bibitem{ref10_3}
C. Yadav and B. S. Reniwal, "Highly Accurate and Energy Efficient Convolutional Spiking Neural Network for Traffic Sign Recognition," \textit{IEEE Access}, vol. 13, pp. 212341–212354, 2025.

\bibitem{ref_DALIF}
T. Zhang, K. Yu, J. Zhang, and H. Wang, "Da-LIF: Dual adaptive leaky integrate-and-fire model for deep spiking neural networks," in \textit{2025 IEEE Int. Conf. Acoust., Speech Signal Process. (ICASSP)}, pp. 1–5, 2025.

\bibitem{ref11}
G. Li, L. Deng, H. Tang, et al., "Brain-inspired computing: A systematic survey and future trends," \textit{Proc. IEEE}, vol. 112, no. 6, pp. 544–584, 2024.

\bibitem{ref12}
Y. Li, S. Deng, X. Dong, R. Gong, and S. Gu, "A free lunch from ANN: Towards efficient, accurate spiking neural networks calibration," in \textit{Proc. Int. Conf. Mach. Learn. (ICML)}, pp. 6316–6325, 2021.

\bibitem{ref13}
J. Kaiser, J. C. V. Tieck, C. Hubschneider, et al., "Towards a framework for end-to-end control of a simulated vehicle with spiking neural networks," in \textit{Proc. 2016 IEEE Int. Conf. Simul. Model. Program. Auton. Robots (SIMPAR)}, pp. 127–134, 2016.

\bibitem{ref14}
Z. Bing, C. Meschede, G. Chen, A. Knoll, and K. Huang, "Indirect and direct training of spiking neural networks for end-to-end control of a lane-keeping vehicle," \textit{Neural Netw.}, vol. 121, pp. 21–36, 2020.

\bibitem{ref15}
P. Chandarana, J. Ou, and R. Zand, "An adaptive sampling and edge detection approach for encoding static images for spiking neural networks," in \textit{Proc. 2021 12th Int. Green Sustain. Comput. Conf. (IGSC)}, pp. 1–8, 2021.

\bibitem{ref16}
Y. Kim and P. Panda, "Revisiting batch normalization for training low-latency deep spiking neural networks from scratch," \textit{Front. Neurosci.}, vol. 15, p. 773954, 2021.

\bibitem{ref17}
H. Zheng, Y. Wu, L. Deng, Y. Hu, and G. Li, "Going deeper with directly-trained larger spiking neural networks," in \textit{Proc. AAAI Conf. Artif. Intell.}, vol. 35, no. 12, pp. 11062–11070, 2021.

\bibitem{ref18}
S. Deng, Y. Li, S. Zhang, and S. Gu, "Temporal efficient training of spiking neural network via gradient re-weighting," \textit{arXiv:2202.11946}, 2022.

\bibitem{ref19}
H. Chen, Y. Liu, W. Ye, and C. Han, "Traffic sign recognition model based on spiking neural network," in \textit{Proc. 2024 4th Int. Conf. Neural Netw. Inf. Commun. Eng. (NNICE)}, pp. 693–696, 2024.

\bibitem{ref_Spikformer}
Z. Zhou, Y. Zhu, C. He, Y. Wang, S. Yan, Y. Tian, and L. Yuan, "Spikformer: When spiking neural network meets transformer," \textit{arXiv:2209.15425, 2022}.

\bibitem{ref_MSResNet}
Y. Hu, L. Deng, Y. Wu, M. Yao, and G. Li, "Advancing spiking neural networks toward deep residual learning," \textit{IEEE Trans. Neural Netw. Learn. Syst.}, vol. 36, no. 2, pp. 2353–2367, 2024.

\bibitem{ref_Shrink}
Y. Ding, L. Zuo, M. Jing, P. He, and Y. Xiao, "Shrinking your timestep: Towards low-latency neuromorphic object recognition with spiking neural networks," in \textit{Proc. AAAI Conf. Artif. Intell.}, vol. 38, no. 10, pp. 11811–11819, 2024.

\bibitem{ref20}
K. Yu, T. Zhang, Q. Xu, G. Pan, and H. Wang, "TS-SNN: Temporal shift module for spiking neural networks," \textit{arXiv:2505.04165}, 2025.

\bibitem{ref_yadav}
C. Yadav, and B. S. Reniwal, “Highly Accurate and Energy Efficient Convolutional Spiking Neural Network for Traffic Sign Recognition,” \textit{IEEE Access}, vol. 13, pp. 212341–212354, 2025.

\bibitem{ref_Deploy}
S. Shen, J. Yang, H. Zhong, H. Lu, X. Zheng, and H. Yang, "Deployment-friendly lane-changing intention prediction powered by brain-inspired spiking neural networks," in \textit{2025 IEEE Intell. Veh. Symp. (IV)}, pp. 2310–2316, 2025.

\bibitem{ref21}
S. C. Kak, "Quantum neural computing," \textit{Adv. Imaging Electron Phys.}, vol. 94, pp. 259–313, 1995.

\bibitem{ref_ANN2SNN}
Z. Hao, T. Bu, J. Ding, T. Huang, and Z. Yu, "Reducing ANN–SNN conversion error through residual membrane potential," in \textit{Proc. AAAI Conf. Artif. Intell.}, vol. 37, no. 1, pp. 11–21, 2023.

\bibitem{ref_SEW}
W. Fang, Z. Yu, Y. Chen, T. Huang, T. Masquelier, and Y. Tian, "Deep residual learning in spiking neural networks," in \textit{Adv. Neural Inf. Process. Syst.}, vol. 34, pp. 21056–21069, 2021.

\bibitem{ref24}
N. Wiebe, A. Kapoor, and K. M. Svore, "Quantum algorithms for nearest-neighbor methods for supervised and unsupervised learning," \textit{Quan. Inf. Com.}, vol. 15, nos. 3–4, pp. 316–356, 2015.

\bibitem{ref26}
J. Preskill, "Quantum computing in the NISQ era and beyond," \textit{Quantum}, vol. 2, p. 79, 2018.

\bibitem{ref27}
E. Farhi and H. Neven, "Classification with quantum neural networks on near-term processors," \textit{arXiv:1802.06002}, 2018.

\bibitem{ref28}
M. Schuld, A. Bocharov, K. M. Svore, and N. Wiebe, "Circuit-centric quantum classifiers," \textit{Phys. Rev. A}, vol. 101, no. 3, p. 032308, 2020.

\bibitem{ref29}
Z. Qu, Z. Chen, S. Dehdashti, et al., "QFSM: A novel quantum federated learning algorithm for speech emotion recognition with minimal gated unit in 5G IoV," \textit{IEEE Trans. Intell. Veh.}, vol. 9, no. 10, pp. 6512–6523, 2024.

\bibitem{ref30}
Z. Qu, X. Zhao, L. Sun, and G. Muhammad, “DAQFL: Dynamic aggregation quantum federated learning algorithm for intelligent diagnosis in Internet of Medical Things,” \textit{IEEE Internet Things J.}, vol. 12, no. 19, pp. 39313–39325, 2025.

\bibitem{ref31}
D. Brand and F. Petruccione, "A quantum leaky integrate-and-fire spiking neuron and network," \textit{npj Quant. Inf.}, vol. 10, no. 1, pp. 1–8, 2024.

\bibitem{ref_jha}
R. K. Jha, N. Kasabov, S. Bhattacharyya, et al., “A hybrid spiking neural network-quantum framework for spatio-temporal data classification: a case study on EEG data,” \textit{EPJ Quantum Technol.}, vol. 12, no. 1, pp. 1–23, 2025.

\bibitem{ref32}
S. Liu and Y. Gu, "Quantum spiking neural networks for image classification," in \textit{Proc. 3th Int. Conf. Alg., Network Com. Tech.(ICANCT)}, vol. 13545, pp. 181–188, 2025.

\bibitem{ref33}
Y. Wu, L. Deng, G. Li, J. Zhu, and L. Shi, "Spatio-temporal backpropagation for training high-performance spiking neural networks," \textit{Front. Neurosci.}, vol. 12, p. 331, 2018.

\bibitem{ref34}
M. Schuld, V. Bergholm, C. Gogolin, J. Izaac, and N. Killoran, "Evaluating analytic gradients on quantum hardware," \textit{Phys. Rev. A}, vol. 99, no. 3, p. 032331, 2019.

\bibitem{ref35}
J. Stallkamp, M. Schlipsing, J. Salmen, and C. Igel, "The German traffic sign recognition benchmark: A multi-class classification competition," in \textit{Proc. Int. Joint Conf. Neural Netw.}, pp. 1453–1460, 2011.

\bibitem{ref36}
National Natural Science Foundation of China, "Chinese traffic sign database,"[Online]. \textit{https://nlpr.ia.ac.cn/pal/trafficdata/recognition.html}.

\bibitem{ref37}
M. Yao, J. Hu, Z. Zhou, L. Yuan, Y. Tian, B. Xu, and G. Li, "Spike-driven transformer," in \textit{Adv. Neural Inf. Process. Syst.}, vol. 36, pp. 64043–64058, 2023.

\bibitem{ref38}
N. Killoran, J. Izaac, N. Quesada, et al., "Strawberry fields: A software platform for photonic quantum computing," \textit{Quantum}, vol. 3, p. 129, 2019.

\bibitem{ref39}
M. A. Nielsen and I. L. Chuang, "Quantum computation and quantum information," \textit{Quantum Inf. Comput.}, 2010.

\end{thebibliography}
\end{document}